\documentclass{article}

    \usepackage[final]{neurips_2023}

\usepackage[utf8]{inputenc} %
\usepackage[T1]{fontenc}    %
\usepackage[pagebackref,breaklinks,colorlinks]{hyperref}
\usepackage{url}            %
\usepackage{booktabs}       %
\usepackage{amsfonts}       %
\usepackage{nicefrac}       %
\usepackage{microtype}      %
\usepackage{xcolor}         %
\usepackage{lmodern}

\usepackage{wrapfig}
\usepackage{amssymb}
\usepackage{stmaryrd}
\usepackage{xfrac}
\usepackage{float}
\usepackage{multirow}
\usepackage{amsmath}
\usepackage{subcaption}
\usepackage{bm}
\usepackage{enumitem}
\usepackage{multicol}
\usepackage[
  ruled,
  vlined,
  linesnumbered,
]{algorithm2e}

\SetCommentSty{algcommentfont}
\usepackage{cleveref}
\crefformat{footnote}{#2\footnotemark[#1]#3}

\usepackage{minitoc}

\newcommand{\ra}[1]{\renewcommand{\arraystretch}{#1}}

\newcommand{\explainerStyle}[1]{{\small\texttt{#1}}}
\newcommand{\SHAP}{\explainerStyle{SHAP}}
\newcommand{\SHAPR}{\explainerStyle{SHAPR}}
\newcommand{\LIME}{\explainerStyle{LIME}}
\newcommand{\PDP}{\explainerStyle{PDP}}
\newcommand{\MAPLE}{\explainerStyle{MAPLE}}

\newcommand{\eg}{e.g.}
\newcommand{\ie}{i.e.}

\newcommand{\figref}[1]{Figure~\ref{#1}}

\title{How Well Do Feature-Additive Explainers Explain Feature-Additive Predictors?}

\author{%
  Zachariah Carmichael\thanks{Code available at \href{https://github.com/craymichael/PostHocExplainerEvaluation}{github.com/craymichael/PostHocExplainerEvaluation}} \quad Walter J. Scheirer \\
  Department of Computer Science \& Engineering\\
  University of Notre Dame\\
  \texttt{\{zcarmich,walter.scheirer\}@nd.edu} \\
}

\makeatletter
\renewcommand\paragraph{\@startsection{paragraph}{4}{\z@}%
  {0.375ex \@plus0.125ex \@minus0.05ex}%
  {-1em}%
  {\normalfont\normalsize\bfseries}}
\makeatother

\begin{document}

\maketitle

\begin{abstract}
  Surging interest in deep learning from high-stakes domains has precipitated concern over the inscrutable nature of black box neural networks. Explainable AI~(XAI) research has led to an abundance of explanation algorithms for these black boxes. Such post hoc explainers produce human-comprehensible explanations, however, their fidelity with respect to the model is not well understood -- explanation evaluation remains one of the most challenging issues in XAI. In this paper, we ask a targeted but important question: can popular feature-additive explainers (\eg{}, \LIME{}, \SHAP{}, \SHAPR{}, \MAPLE{}, and \PDP{}) explain feature-additive predictors? Herein, we evaluate such explainers on ground truth that is analytically derived from the additive structure of a model. We demonstrate the efficacy of our approach in understanding these explainers applied to symbolic expressions, neural networks, and generalized additive models on thousands of synthetic and several real-world tasks. Our results suggest that all explainers eventually fail to correctly attribute the importance of features, especially when a decision-making process involves feature interactions.
\end{abstract}

\section{Introduction}\label{sec:introduction}

The counterintuitive mispredictions and undesirable behaviors of black box AI systems~\cite{quionero-candelaDatasetShiftMachine2009,DBLP:journals/corr/SzegedyZSBEGF13,hendrycksNaturalAdversarialExamples2019,adversarialExamples,liptonMythos2018} has piqued widespread interest in explainable AI (XAI) solutions, including from the medical, financial, and the legal domains~\cite{liptonMythos2018,millerAIMedical2018,Rudin2019,medicalXAIsurvey2020}.
Late interest has saturated due to real-world consequences~\cite{oneilWeaponsMathDestruction2016,buolamwiniGenderShadesIntersectional2018,Bommasani2021FoundationModels,incidentDB}
and regulatory pushes~\cite{EU-GDPR,EU-US-TTC_statement2021,EU-AI-Act,US-Alg-Acct-Act,SchumerSAFE}.
Accordingly, a plethora of XAI approaches have been proposed to shed light on previously inscrutable black boxes. However, explanations are notoriously difficult to evaluate~\cite{doshi2017towards}. Measuring the fidelity of explanations has remained so unverifiable~\cite{bhalla2023verifiable} that we are starting to see meta-evaluations (quality evaluations of quality evaluation metrics of explanations of black box models)~\cite{hedstrom2023meta}. 

In this work, we study the evaluation of a specific but popular class of XAI methods: post hoc feature-additive explainers, like \LIME{}~\cite{ribeiroWhyShouldTrust2016}, \SHAP{}~\cite{lundbergUnifiedApproachInterpreting2017}, and \PDP{}~\cite{Friedman2001}.
We ask, ``can feature-additive explainers explain feature-additive predictors?''
We propose a novel explainer evaluation methodology that overcomes many issues present in prior work.
This work presents the following contributions:
\begin{itemize}[noitemsep,nolistsep,leftmargin=*]
    \item We construct a test bed for the evaluation of feature-additive post hoc explanations against \textit{ground truth} derived analytically from feature-additive models. By definition, perfect explanations should be \emph{exactly equal} to this ground truth.
    \item To facilitate evaluation, we propose an algorithm, \textsc{MatchEffects}, that directly maps any model with any amount of additive structure to feature-additive post hoc explanations.
    \item We evaluate the popular post hoc explainers \LIME{}~\cite{ribeiroWhyShouldTrust2016}, \SHAP{}~\cite{lundbergUnifiedApproachInterpreting2017}, \SHAPR{}~\cite{aasExplainingIndividualPredictions2020}, \PDP{}~\cite{Friedman2001}, and \MAPLE{}~\cite{plumbModelAgnosticSupervised2018}
    on thousands of synthetic tasks and models, as well as with
    neural networks and generalized additive models
    on several real-world datasets.
    \item We demonstrate that although \SHAP{}
    outperforms the other explainers, all explainers begin to fail in the presence of higher-dimensional data,  models with higher-order interactions, and models with more interaction effects.
\end{itemize}

\section{Background}\label{sec:background}

\paragraph{Algorithms for Local Post Hoc Explanation}\label{sec:posthoc-background}

Whereas \textit{ante hoc} explainers have an intrinsic notion of interpretability, post hoc methods serve as a surrogate explainer for a black box.
There are several classes of post hoc explanation methods, including salience maps~\cite{bachPixelWiseExplanationsNonLinear2015,selvarajuGradCAMVisualExplanations2017},
local surrogate models~\cite{lundbergUnifiedApproachInterpreting2017,ribeiroWhyShouldTrust2016},
counterfactuals~\cite{DBLP:conf/ecai/WhiteG20},
and global interpretation techniques~\cite{zhangInterpretingCNNsDecision2019}.
A comprehensive overview can be found in~\cite{guidottiSurveyMethodsExplaining2018,molnar2019,DBLP:journals/corr/abs-2010-10596,Schwalbe2023,ALI2023101805}.
However, here we strictly focus on feature-additive local approximation~\cite{camburuCanTrustExplainer2019}, which is one of the most prevalent explanation strategies.
Local post hoc explainers
estimate the feature importance for a single decision whereas
global explainers provide explanations of a model for an entire dataset.
Explainers aim to recover the local model response about an instance while isolating the most important features to produce comprehensible explanations.
Specifically, we consider the \LIME{}~\cite{ribeiroWhyShouldTrust2016}, \SHAP{}~\cite{lundbergUnifiedApproachInterpreting2017}, \SHAPR{}~\cite{aasExplainingIndividualPredictions2020}, \PDP{}~\cite{Friedman2001}, and \MAPLE{}~\cite{plumbModelAgnosticSupervised2018} explainers. The explainers are detailed in Appendix A.

\paragraph{Evaluation of Explainers}\label{sec:posthoc-eval-background}

There are three main types of evaluations: application-grounded (real humans, real tasks), human-grounded (real humans, simplified tasks), and functionally-grounded (no humans, proxy tasks)~\cite{doshi2017towards}.
Human- and application-grounded evaluations are expensive, subjective, and qualitative. However, they measure the human utility and effectiveness of explanations. Functionally-grounded metrics are concerned with proxies for the same objectives, but also can quantitatively score the fidelity of an explanation with respect to the model being explained~\cite{anecdotalEvidenceXAI2023}.
We are interested in the \textit{functionally-grounded} evaluation of explanation \textit{fidelity} (correctness) in this paper.
The highest-fidelity evaluations involve comparing explanations to the ground truth explanation. For the sake of space, we abbreviate related work and elaborate in Appendix A. In short, there are three types of ground truth checks that can be performed: against annotations, controlled data, and white boxes~\cite{info14080469,anecdotalEvidenceXAI2023}. While annotation-based and controlled data checks involve subjective or proxy definitions of ground truth, white box checks do not as the explanation is implicit in the form of the model. Specifically, we care about evaluating \textit{exact} feature contributions in explanations rather than just feature selection or ranking.

Our approach has several advantages over prior work.
(1) \cite{NEURIPS2020_ce758408,pmlr-v119-lakkaraju20a} define the exact feature contributions as the coefficients of a linear regression model to evaluate the fidelity of explainers, such as \LIME{} and \SHAP{}. However, this is inappropriate for these explainers and thus does not provide an exact set of feature contributions (see Appendix A for details). We correct for these issues in our work for all considered explainers.
(2) No prior work on white box checks has considered the case of feature interactions~\cite{miro2023novel,NEURIPS2020_ce758408,pmlr-v119-lakkaraju20a}, which are ubiquitous among black box models, especially neural networks. We consider models with a various number of feature interactions and order of feature interactions.
(3) Unlike prior work~\cite{miro2023novel,NEURIPS2020_ce758408,pmlr-v119-lakkaraju20a}, we consider in our experiments both synthetic and real-world data, both tabular and image data, as well as both non-learnable and learnable models, including convolutional neural networks.

Further comprehensive overviews of explanation evaluation methodologies and aspects
are detailed in~\cite{info14080469,anecdotalEvidenceXAI2023,zhouEvaluatingQualityMachine2021,Vermeire2022,pawelczyk2021carla,DBLP:journals/adac/RamonMPE20} (as well as Appendix A).

\begin{figure*}%
    \centering
    \includegraphics[width=.73\linewidth]{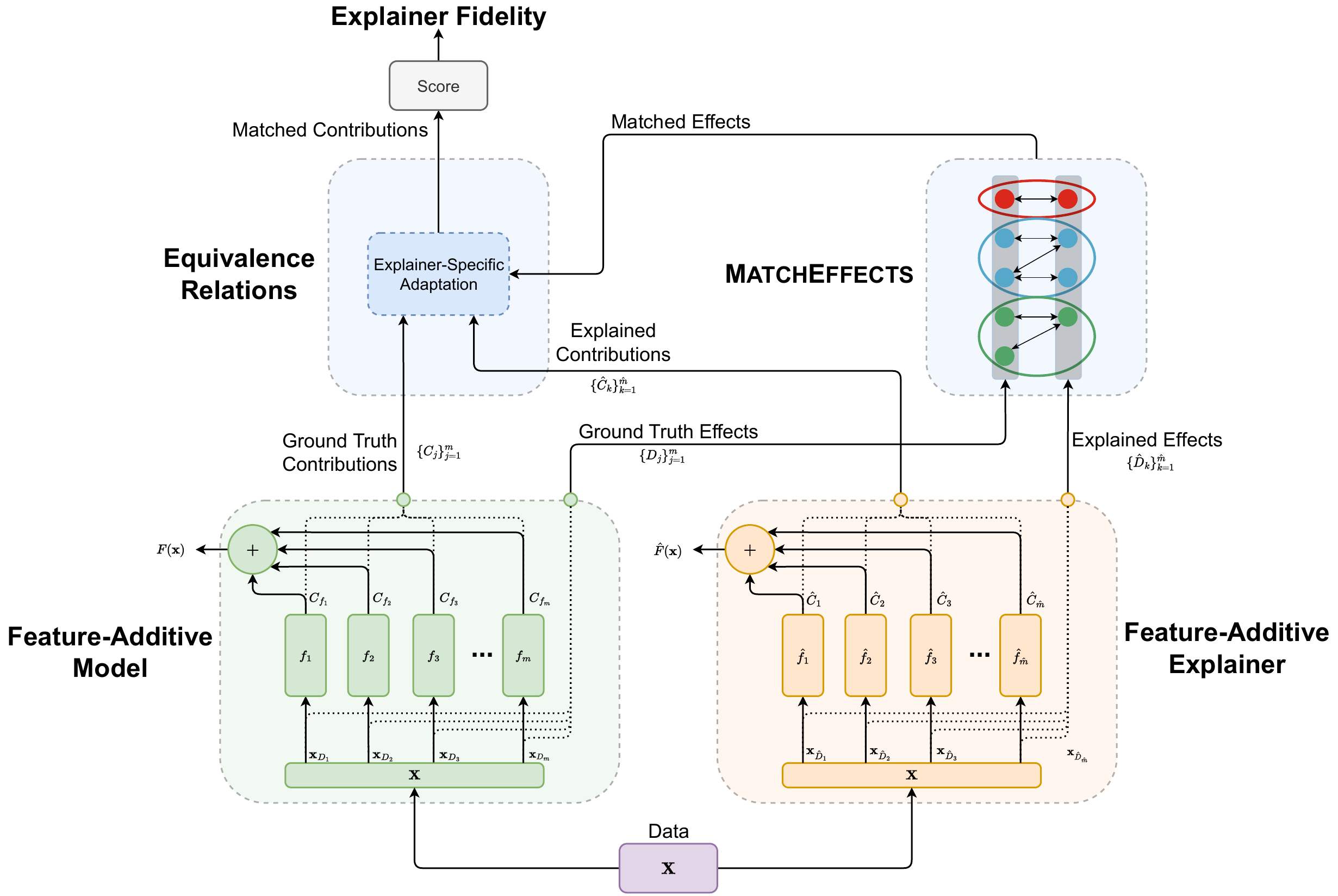}
    \caption{High-level overview of the proposed evaluation of post hoc explainer quality. Therein, a feature-additive post hoc explainer estimates the feature contributions of a feature-additive model for some data $\mathbf{X}$. Each $f_i$ and $\hat{f}_j$ is a function as described in Section~\ref{sec:model-formulation}. Both the model and the explainer produce a set of effects and contributions -- those of the model are the ground truth. To compare the model and explainer, the effects of the explainer are aligned with the ground truth using the \textsc{MatchEffects} algorithm (Section~\ref{lab:gt_align}). Thereafter, the matched effects inform how to carry out the equivalence relations -- this process allows for direct comparison of explained contributions to the ground truth (Section~\ref{sec:equiv-relation}). Finally, the fidelity of the explainer is computed using the matched contributions -- perfect explainer explanations should be \textit{exactly equal} to the ground truth explanations.}
    \label{fig:framework}
\end{figure*}

\section{Methodology}\label{sec:method}

We propose to evaluate feature-additive explainers by comparing their explanations to the ground truth explanations from feature-additive white box models. 
We provide a fair way of comparing these two explanations (sets of feature contributions) when the model has feature interactions, or, more generally, when the explainer and model explanations comprise different sets of effects.
\figref{fig:framework} shows a high-level overview of our approach.

\paragraph{Feature-Additive Model \& Explainer Formulation}\label{sec:model-formulation}
We consider a general form of feature-additive white box models, similar to, but distinct from, generalized additive models (GAMs)\footnote{This formulation notably differs from GAMs in that each $f_j(\cdot)$ can be non-smooth.}.
Concise definitions of feature contributions follow naturally from its additive structure while still allowing for feature interactions, high dimensionality, and highly nonlinear effects.
Let $\mathbf{X} \in \mathbb{R}^{n \times d}$ be a matrix with $n$ samples and $d$ features,
$\mathbf{x} \in \mathbf{X}$ be a sample,
$D = \{i\}_{i=1}^{d}$
be the set of all feature indices, and $F(\cdot)$ be an additive function comprising $m$ effects. Each effect is given by a non-additive function $f_j(\cdot)$
\begin{wrapfigure}{r}{0.4\textwidth}
\vspace{-1.5ex}
\begin{align}\label{eq:additive_model}
    F(\mathbf{x}) &= \sum_{j=1}^{m} f_j\left(\mathbf{x}_{D_{j}}\right) = \sum_{j=1}^{m} C_{j} \\
\label{eq:additive_explainer}
    \hat{F}(\mathbf{x}) &= 
    \sum_{k=1}^{{\hat{m}}} \hat{f}_k\left(\mathbf{x}_{\hat{D}_{k}}\right) = 
    \sum_{k=1}^{{\hat{m}}} \hat{C}_{k}
\end{align}
\vspace{-1.5ex}
\end{wrapfigure}
that takes a subset of features $D_{j} {\subseteq} D$ as input and yields an additive contribution $C_{j}$ to
the model output as shown in Eq.~\eqref{eq:additive_model}.
In this paper, we refer to an \textit{effect} by the subset of features $D_{j}$ that it comprises. If $|D_{j}|>1$, it is an interaction effect, otherwise it is a main effect.
The ground truth explanation is then the set of effects and their contributions
$\{(D_{j}, C_{j})\}_{j=1}^{m}$.

\textit{Explainer}~
For explainers, we denote local estimates of the model as $\hat{F}(\cdot)$ which comprise a summation of ${\hat{m}}$ effects given by each $\hat{f}_k(\cdot)$ as in Eq.~\eqref{eq:additive_explainer}.
Similarly, the explanation from an explainer has the form $\{(\hat{D}_{k}, \hat{C}_{k})\}_{k=1}^{\hat{m}}$.
The explainers evaluated in this work have $\hat{m} \le d$, however, explainers with $\hat{m} > d$ are compatible with our formulation and implementation.

\textit{Synthetic Models}~
We generate synthetic models with controlled degrees of sparsity, order of interaction, nonlinearity, and size. This allows us to study how different model characteristics affect explanation quality. Here, each $f_j(\cdot)$ is a composition of random non-additive unary and/or binary operators for a random subset of features $D_{j}$.
Expressions are generated based on these parameters and we verify that the domains and ranges are in $\mathbb{R}$.
For example, a generated expression with $m = d = 4$ and 2 dummy features could look like $F(\mathbf{x}) = \mathbf{x}_{1} + e^{\mathbf{x}_{4}} +
\log(\mathbf{x}_{1} \mathbf{x}_{4})+ \frac{\mathbf{x}_{4}}{\mathbf{x}_{1}}$.
See
Appendix E
for details of our algorithm used to generate such models.

\textit{Learned Models}~
We consider two types of learned models: GAMs and feature-additive neural networks~(NNs).
The former is a rich yet simple model that models nonlinear effects while being conducive for understanding feature significance~\cite{DBLP:books/lib/HastieTF09}. Each $f_j(\cdot)$ is a smooth nonparametric function that is fit using splines. A link function relates the summation of each $f_j(\cdot)$ to the target response, such as the identity link for regression and the logit link for classification.

The feature-additive NNs we consider have the same additive structure, but each $f_j(\cdot)$ is instead a fully-connected NN. Each NN can have any architecture, operates on $D_{j}$, and yields a scalar value for regression or a vector for classification. The output is the summation of each NN with a link function similar to the GAM.
This structure is related to the neural additive model proposed in~\cite{agarwalNeuralAdditiveModels2020}.
This NN formulation also holds for convolutional NNs~(CNNs), which can have a non-unary $m$ as long as the receptive field at any layer does not cover the full image.
Notably, while the CNN operates on the image data $\bm{\mathcal{X}}\in\mathbb{R}^{n\times d_1\times d_2 \times d_3}$, the explainers that we consider operate on the flattened data $\mathbf{X}\in\mathbb{R}^{n\times d_1 d_2 d_3}$.
See Appendix B for more details.

The number of effects $m$ and each effect $D_{j}$ are selected randomly for learned models such that $m > 1$, the number of matches from \textsc{MatchEffects} (introduced in Section~\ref{lab:gt_align}) is ${>}1$, and the task error is satisfactorily low.

\paragraph{Ground Truth Alignment: \textsc{MatchEffects}}\label{lab:gt_align}

With our formalism, we now have a model and an explainer, each of which produces explanations as a set of effects and their corresponding contributions.
Because there may not be a one-to-one correspondence between the two sets,
we cannot directly compare the effects.
Consider the case of a model with an interaction effect, \textit{\ie{}}, some $|D_{j}| \ge 2$; if the explanation has no $\hat{D}_{k} = D_{j}$, then a direct comparison of explanations is not possible.
To this end, we propose the \textsc{MatchEffects} algorithm, 
which matches subsets of effects
between the model and explainer.
Put simply, the algorithm finds the smallest feature interaction effects that are common between the model and explainer explanations. For example, if an explainer explanation contains contributions for features 1 and 2, and the model ground truth explanation contains a contribution for the interaction effect involving both 1 and 2, then the sum of the explainer contributions for these features is compared to the ground truth contribution.

To achieve this matching, we consider all $D_{j}$ and $\hat{D}_{{f}_k}$ to be the left- and right-hand vertices, respectively, of an undirected bipartite graph.
Edges are added between effects with common features.
We then find the connected components of this graph to identify groups of effects with inter-effect dependencies.
If every component contains an exact match, for example, if $\text{match}_F {=} \{\{2\}, \{2, 3\}\}$ and $\text{match}_{\hat{F}} {=} \{\{2\}, \{2, 3\}\}$, then each contribution by $\{2\}$ and $\{2, 3\}$ will be compared separately.
Further details and algorithm illustrations are provided in Appendix A.

\paragraph{Equivalence Relations to Explainers}\label{sec:equiv-relation}
\begin{wrapfigure}{r}{0.26\textwidth}
\vspace{-1.5ex}
\begin{align}%
\label{eq:unnorm1}
\theta_0' &= \theta_0 - \sum_i \frac{\mu_i \theta_i}{\sigma_i}\\
\label{eq:unnorm2}
\theta_i' &= \frac{\theta_i}{\sigma_i}%
\end{align}%
\vspace{-1.5ex}
\end{wrapfigure}
With \textsc{MatchEffects} and MaIoU defined, a direct comparison between
true and explained explanations is nearly possible.
However, some adaptation is still required due to the use of normalization and differing definitions of ``contribution'' between explainers.
Here, we bridge together these definitions.
\LIME{} normalizes the data as z-scores, \textit{\ie{}}, $z = (x_i - \mu_i) / \sigma_i$, before learning
a linear model. We then need to scale the coefficients $\Theta = \{\theta_i\}_{i=1}^{d}$ of each local linear model using the estimated means $\mu_i$ and standard deviations $\sigma_i$ from the data as in Eqs.~\eqref{eq:unnorm1} and \eqref{eq:unnorm2}.
In \SHAP{}, the notion of feature importance is the approximation of the mean-centered independent feature contributions for an instance. The expected value $\mathbb{E}[F(\mathbf{x})]$ is estimated from the background data\, \SHAP{}\, receives.\,
In\, order\, to\, allow\, for
\vspace{-1.1ex}\WFclear%
\begin{wrapfigure}{r}{0.55\textwidth}
\vspace{-1.5ex}
\begin{align}\label{eq:shapunnorm}
    C_{\text{match}_{\hat{F}}} &= \sum_{k \in \text{match}_{\hat{F}}} \hat{f}_k(\mathbf{x}_{k}) + \sum_{j \in \text{match}_{F}} \mathbb{E}[C_{j}]
\end{align}
\vspace{-1.5ex}
\end{wrapfigure}
valid comparison, we add back the expected value of the true contribution
$\mathbb{E}[C_{i}]$ estimated from the same data. However, since a 1:1 matching is not a guarantee, we must consider all effects grouped by said matching as in Eq.~\eqref{eq:shapunnorm}.
The same procedure applies to \SHAPR{}.
See
Appendix C
for the derivations of these relations.
Furthermore, \LIME{} and \MAPLE{} provide feature-wise explanations as the coefficients $\Theta$ of a linear regression model. In turn, we must simply compute the product between each coefficient and feature vector $\mathbf{x}_{i} \,\theta_i$ to yield the contribution to the output according to the explainer.

\section{Experimental Results}\label{sec:results}
\renewcommand{\thempfootnote}{\arabic{mpfootnote}}
\begin{figure*}%
    \centering
    \begin{minipage}{\linewidth}
    \begin{subfigure}{.9\linewidth}
        \includegraphics[width=\linewidth]{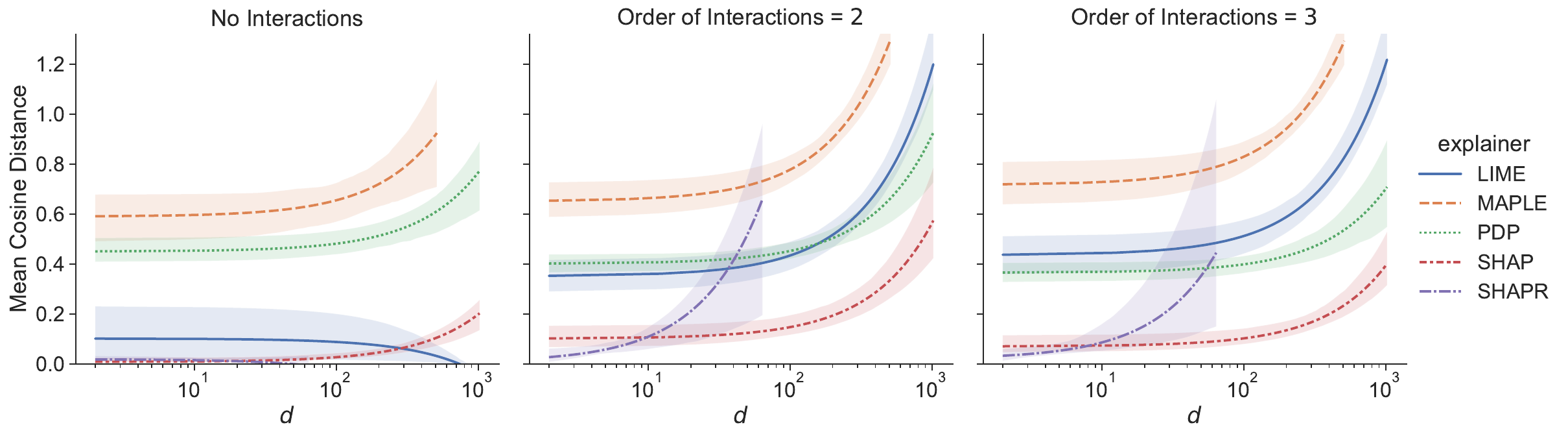}
        \label{fig:synthetic_1a}
      \end{subfigure}\\%
      \begin{subfigure}{.9\linewidth}
        \includegraphics[width=\linewidth]{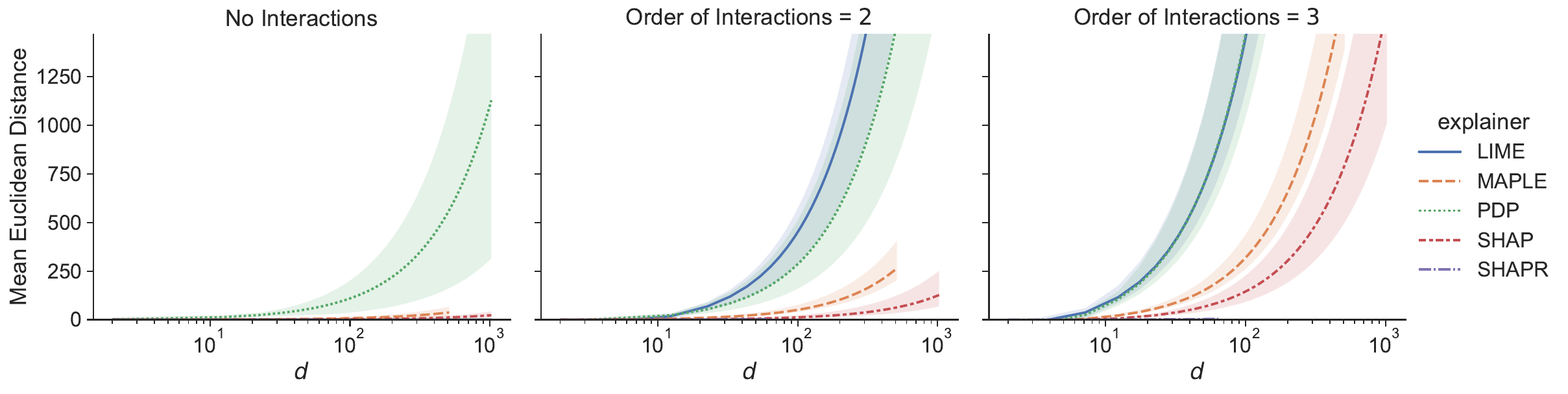}
        \label{fig:synthetic_1b}
      \end{subfigure}%
    \caption{Average cosine distances (top) and Euclidean distances (bottom) between ground truth and explained effect contributions as a function of the number of features $d$ and the order of interaction effects.
    As the dimensionality, the degree of interactions, and the number of interactions increase, the disagreement with ground truth grows for all explainers.
    }
    \label{fig:synthetic_1}
    \end{minipage}
\end{figure*}
We evaluate the explainers on thousands of synthetic problems and popular real-world datasets.
By varying the data and models, we identify when explainers fail, whether plausible explanations are faithful, and other interesting trends.
See Appendix B for experimental setup and implementation details.

\paragraph{Evaluation}\label{sec:evaluation}
We measure the error between the explanations of the ground truth and explainer using a few metrics.
The set of contributions comprising each explanation can be thought of as a vector collectively, thus we can compute the distance between them after applying \textsc{MatchEffects}. First considered is Euclidean distance to understand the magnitude of the disagreement with the ground truth. To quantify the disagreement in orientation, we utilize cosine distance.
Each measure of error here is the quantified \textit{infidelity} of explanations.

\paragraph{Synthetic Problems}\label{sec:synth-problems}
We first demonstrate our approach on 2,000 synthetic models that are generated with a varied number of effects, order of interaction, number of features, degree of nonlinearity, and number of unused (dummy) variables. The explainers are evaluated on each model with access to the full dataset and black box access to the model. Some explainers failed to explain some models -- see Appendix B for details.

Results demonstrate the efficacy of the proposed approach in understanding explanation quality, as well as factors that influence it when paired with the experimental design.
As the dimensionality, the degree of interactions, and the number of interactions increase, the disagreement between ground truth and explanation grows.
\figref{fig:synthetic_1} illustrates these results for all of the explained synthetic models.
Because \LIME{} failed to explain a substantial portion of synthetic models, it appears to improve with an increased $d$ in the leftmost plots;
in reality, it only succeeded in explaining simpler models with a larger $d$.
\SHAP{} performs the best relative to the other explainers, maintaining both a closer and more correctly-oriented explanation compared to the ground truth.
Interestingly, the ranking of \LIME{} and \MAPLE{} swaps when comparing average cosine and Euclidean distances.
Surprisingly, \SHAPR{} struggles to handle interaction effects effectively -- the baseline \SHAP{} outperforms it substantially.
Appendix D
includes additional analyses and figures with synthetic models, including evaluation as a function of the number of number of interactions, number of nonlinearities, and dummy features.

\begin{table*}%
    \small
    \centering
    \ra{1.1}
    \begin{tabular}{@{}lccccccccc@{}}
        \toprule
        \multirow{2}{*}{Dataset} & \multirow{2}{*}{Model} && \multicolumn{5}{c}{Explainer Error} && \multirow{2}{*}{$\rho_{\text{perf}}$}\\
        \cmidrule{4-8}
        &&& {\PDP{}} & {\LIME{}} & {\MAPLE{}} & {\SHAP{}} & {\SHAPR{}} \\
        \midrule
        \multirow{2}{*}{Boston}
            & GAM && 0.340 & 0.709 & 0.652 & 0.001 & 0.111 && 0.995 \\
            &  NN && 0.278 & 0.182 & 0.431 & 0.001 & 0.209 && 0.351 \\[.9ex]
        \multirow{2}{*}{COMPAS}
            & GAM && 0.821 & 0.781 & 0.863 & 0.000 & -- && 0.800 \\
            &  NN && 0.328 & 0.062 & 0.274 & 0.001 & -- && 1.000 \\[.9ex]
        \multirow{2}{*}{FICO}
            & GAM && 0.795 & 0.949 & 0.962 & 0.003 & -- && 0.200 \\
            &  NN && 0.761 & 0.193 & 0.270 & 0.001 & -- && 0.800 \\[.9ex]
        \multirow{1}{*}{MNIST}
            & CNN && 0.660 & 0.253 & 0.318 & 0.049 & 0.175 && 0.410 \\
        \bottomrule
    \end{tabular}
    \caption{Real-world explainer results on several datasets for GAMs and NNs. Here, explainer error is the cosine distance averaged over all samples and classes, if applicable. $\rho_{\text{perf}}$ is Spearman's rank correlation coefficient between the mean explanation cosine distance and explainer accuracy. \SHAPR{} is not implemented for data with categorical variables in this work.}
    \label{tab:real_world_results}
\end{table*}

\paragraph{Real-World Case Studies}\label{sec:real-problems}

We evaluate GAMs and feature-additive NNs on several real-world datasets: Boston housing~\cite{HARRISON197881}, COMPAS~\cite{angwin2016machine}, FICO HELOC~\cite{ficoheloc2018}, and a down-sampled version of MNIST~\cite{MNIST}.
Table~\ref{tab:real_world_results} contains the aggregate results across all real-world datasets for the considered models. Among the considered explainers, \SHAP{} outperforms on all datasets and models, often by several orders of magnitude.
Surprisingly, \SHAPR{} performs worse than \SHAP{}, but still ranks well compared to the other explainers. \PDP{}, \LIME{}, and \MAPLE{} produce poor explanations in general, and all explainers struggled more with the GAMs than the considered NNs.
To test whether explainer fidelity correlates with accuracy, we compute the Spearman's rank correlation coefficient $\rho_{\text{perf}}$ between the mean explanation cosine similarity (explanation fidelity) and explainer accuracy. Recall that under the feature-additive perspective that the sum of the contributions from an explainer approximates the model output, which can be treated as the prediction of the explainer. The scores, shown in Table~\ref{tab:real_world_results}, demonstrate that a plausible explainer, \textit{i.e}, one that predicts accurately, does not necessarily produce faithful explanations, and vice versa.

\section{Discussion}\label{sec:discussion}

The answer to our question -- whether feature-additive explainers effectively explain feature-additive predictors -- is a nuanced ``no.'' It depends on the application, the data, and the model. However, typical NNs contain a greater number of interaction effects and order of interactions than those considered here -- if an explainer underperforms on feature-additive white boxes, then it should not be expected to perform well with black box predictors.

The shortcomings of these explainers arise from their underlying assumptions, such as feature independence and the locality of linearity.
These assumptions are further impacted by the explainer hyperparameters
which require tuning dependent upon the data and model. In practice, these knobs can be adjusted until the explanations ``look right,'' which is not realistic when the most faithful hyperparameters need to be derived from the black box itself.
This is especially troubling as studies show that data scientists overtrust or do not understand interpretability techniques~\cite{kaurInterpretingInterpretabilityUnderstanding2020,krishna2022disagreement}. With the results of our study, even those practitioners who do not abuse these explanation tools may still be mislead.

Our results corroborate findings in prior research.
Post hoc explainers have been show to be unverifiable, unfaithful, inconsistent, incomplete, intractable, unsuitable for real-time applications, and/or untrustworthy~\cite{swamy2023future,rudin2022black,chaszczewicz2023task,doi:10.1126/science.abg1834,krishna2022disagreement,bordt2022posthoc,DBLP:conf/aaai/BroeckLSS21,carmichaelposthoceval2021,pmlr-v108-garreau20a}.
Additionally, these methods can be fooled~\cite{slackFoolingLIMESHAP2020,dimanov2020you,dombrowski2019explanations,adv_xai_site}.
However, they may increase user trust in AI systems~\cite{carter2022explainable}, user performance under certain conditions~\cite{humer2022comparing}, and trustlessly audit black boxes~\cite{unfooling2023}. Nonetheless, post hoc explanation is often argued to be unsuitable for high-stakes applications~\cite{Rudin2019}.
Rather, intrinsically interpretable models should be favored~\cite{swamy2023future,rudin2022black}, which are more desirable to experts and can even be more accurate than their black box counterparts in high-stakes application domains~\cite{afnan2021interpretable,Jain2022Importance,doi:10.1148/radiol.230574,hatherley2022virtues}.

\emph{Future Work}~
A natural extension of this work would be to evaluate additional explanation methods that consider interaction effects and guide the improvement of explainer quality.
We believe that progress within this class of explainers will emerge by accounting for interdependence between features, better defining locality, and scaling computation for high-dimensional data.
Last, we echo the arguments that XAI research needs to be rigorous with certifiable guarantees, clear and falsifiable hypotheses, and justified generalization checks%
~\cite{geirhos2023don,peters2023unjustified,leavitt2020towards}.

{
\small
\bibliographystyle{plain}
\bibliography{bibby}

\begin{thebibliography}{100}

\bibitem{aasExplainingIndividualPredictions2020}
Kjersti Aas, Martin Jullum, and Anders Løland.
\newblock Explaining individual predictions when features are dependent: More
  accurate approximations to shapley values.
\newblock {\em Artificial Intelligence}, 298:1--24, September 2021.

\bibitem{tensorflow2015-whitepaper}
Mart\'{\i}n Abadi, Ashish Agarwal, Paul Barham, Eugene Brevdo, Zhifeng Chen,
  Craig Citro, Greg~S. Corrado, Andy Davis, Jeffrey Dean, Matthieu Devin,
  Sanjay Ghemawat, Ian Goodfellow, Andrew Harp, Geoffrey Irving, Michael Isard,
  Yangqing Jia, Rafal Jozefowicz, Lukasz Kaiser, Manjunath Kudlur, Josh
  Levenberg, Dan Man\'{e}, Rajat Monga, Sherry Moore, Derek Murray, Chris Olah,
  Mike Schuster, Jonathon Shlens, Benoit Steiner, Ilya Sutskever, Kunal Talwar,
  Paul Tucker, Vincent Vanhoucke, Vijay Vasudevan, Fernanda Vi\'{e}gas, Oriol
  Vinyals, Pete Warden, Martin Wattenberg, Martin Wicke, Yuan Yu, and Xiaoqiang
  Zheng.
\newblock {TensorFlow}: Large-scale machine learning on heterogeneous systems,
  2015.
\newblock Software available from tensorflow.org.

\bibitem{NEURIPS2020_075b051e}
Julius Adebayo, Michael Muelly, Ilaria Liccardi, and Been Kim.
\newblock Debugging tests for model explanations.
\newblock In H.~Larochelle, M.~Ranzato, R.~Hadsell, M.F. Balcan, and H.~Lin,
  editors, {\em Advances in Neural Information Processing Systems}, volume~33,
  pages 700--712. Curran Associates, Inc., 2020.

\bibitem{afnan2021interpretable}
Michael Anis~Mihdi Afnan, Yanhe Liu, Vincent Conitzer, Cynthia Rudin, Abhishek
  Mishra, Julian Savulescu, and Masoud Afnan.
\newblock Interpretable, not black-box, artificial intelligence should be used
  for embryo selection, 2021.

\bibitem{agarwal2022openxai}
Chirag Agarwal, Eshika Saxena, Satyapriya Krishna, Martin Pawelczyk, Nari
  Johnson, Isha Puri, Marinka Zitnik, and Himabindu Lakkaraju.
\newblock {OpenXAI}: Towards a transparent evaluation of post hoc model
  explanations.
\newblock {\em arXiv preprint arXiv:2206.11104}, 2022.

\bibitem{agarwalNeuralAdditiveModels2020}
Rishabh Agarwal, Levi Melnick, Nicholas Frosst, Xuezhou Zhang, Ben Lengerich,
  Rich Caruana, and Geoffrey Hinton.
\newblock Neural additive models: Interpretable machine learning with neural
  nets.
\newblock In A.~Beygelzimer, Y.~Dauphin, P.~Liang, and J.~Wortman Vaughan,
  editors, {\em Advances in Neural Information Processing Systems}, pages
  1--13, 2021.

\bibitem{info14080469}
Nourah Alangari, Mohamed El~Bachir~Menai, Hassan Mathkour, and Ibrahim
  Almosallam.
\newblock Exploring evaluation methods for interpretable machine learning: A
  survey.
\newblock {\em Information}, 14(8), 2023.

\bibitem{ALI2023101805}
Sajid Ali, Tamer Abuhmed, Shaker El-Sappagh, Khan Muhammad, Jose~M.
  Alonso-Moral, Roberto Confalonieri, Riccardo Guidotti, Javier {Del Ser},
  Natalia Díaz-Rodríguez, and Francisco Herrera.
\newblock Explainable artificial intelligence ({XAI}): What we know and what is
  left to attain trustworthy artificial intelligence.
\newblock {\em Information Fusion}, 99:101805, 2023.

\bibitem{angwin2016machine}
Julia Angwin, Jeff Larson, Surya Mattu, and Lauren Kirchner.
\newblock Machine bias: there's software used across the country to predict
  future criminals. {A}nd it's biased against blacks., 2016.
\newblock
  \url{https://www.propublica.org/article/machine-bias-risk-assessments-in-criminal-sentencing}.

\bibitem{doi:10.1126/science.abg1834}
Boris Babic, Sara Gerke, Theodoros Evgeniou, and I.~Glenn Cohen.
\newblock Beware explanations from {AI} in health care.
\newblock {\em Science}, 373(6552):284--286, 2021.

\bibitem{bachPixelWiseExplanationsNonLinear2015}
Sebastian Bach, Alexander Binder, Gr{\'e}goire Montavon, Frederick Klauschen,
  Klaus-Robert M{\"u}ller, and Wojciech Samek.
\newblock On pixel-wise explanations for non-linear classifier decisions by
  layer-wise relevance propagation.
\newblock {\em PLOS ONE}, 10(7):e0130140, July 2015.

\bibitem{adv_xai_site}
Hubert Baniecki.
\newblock Adversarial explainable {AI}.
\newblock \url{https://hbaniecki.com/adversarial-explainable-ai/}, 2023.
\newblock Accessed: 2023-01-28.

\bibitem{NEURIPS2020_56f9f889}
Cher Bass, Mariana da~Silva, Carole Sudre, Petru-Daniel Tudosiu, Stephen Smith,
  and Emma Robinson.
\newblock {ICAM}: Interpretable classification via disentangled representations
  and feature attribution mapping.
\newblock In H.~Larochelle, M.~Ranzato, R.~Hadsell, M.F. Balcan, and H.~Lin,
  editors, {\em Advances in Neural Information Processing Systems}, volume~33,
  pages 7697--7709. Curran Associates, Inc., 2020.

\bibitem{Baumgartner_2018_CVPR}
Christian~F. Baumgartner, Lisa~M. Koch, Kerem~Can Tezcan, Jia~Xi Ang, and Ender
  Konukoglu.
\newblock Visual feature attribution using wasserstein {GAN}s.
\newblock In {\em Proceedings of the IEEE Conference on Computer Vision and
  Pattern Recognition (CVPR)}, June 2018.

\bibitem{bhalla2023verifiable}
Usha Bhalla, Suraj Srinivas, and Himabindu Lakkaraju.
\newblock Verifiable feature attributions: A bridge between post hoc
  explainability and inherent interpretability.
\newblock In {\em {ICML} 3rd Workshop on Interpretable Machine Learning in
  Healthcare ({IMLH})}, 2023.

\bibitem{Bommasani2021FoundationModels}
Rishi Bommasani, Drew~A. Hudson, Ehsan Adeli, Russ Altman, Simran Arora, Sydney
  von Arx, Michael~S. Bernstein, Jeannette Bohg, Antoine Bosselut, Emma
  Brunskill, Erik Brynjolfsson, S.~Buch, Dallas Card, Rodrigo Castellon,
  Niladri~S. Chatterji, Annie~S. Chen, Kathleen~A. Creel, Jared Davis, Dora
  Demszky, Chris Donahue, Moussa Doumbouya, Esin Durmus, Stefano Ermon, John
  Etchemendy, Kawin Ethayarajh, Li~Fei-Fei, Chelsea Finn, Trevor Gale,
  Lauren~E. Gillespie, Karan Goel, Noah~D. Goodman, Shelby Grossman, Neel Guha,
  Tatsunori Hashimoto, Peter Henderson, John Hewitt, Daniel~E. Ho, Jenny Hong,
  Kyle Hsu, Jing Huang, Thomas~F. Icard, Saahil Jain, Dan Jurafsky, Pratyusha
  Kalluri, Siddharth Karamcheti, Geoff Keeling, Fereshte Khani, O.~Khattab,
  Pang~Wei Koh, Mark~S. Krass, Ranjay Krishna, Rohith Kuditipudi, Ananya Kumar,
  Faisal Ladhak, Mina Lee, Tony Lee, Jure Leskovec, Isabelle Levent, Xiang~Lisa
  Li, Xuechen Li, Tengyu Ma, Ali Malik, Christopher~D. Manning, Suvir~P.
  Mirchandani, Eric Mitchell, Zanele Munyikwa, Suraj Nair, Avanika Narayan,
  Deepak Narayanan, Benjamin Newman, Allen Nie, Juan~Carlos Niebles, Hamed
  Nilforoshan, J.~F. Nyarko, Giray Ogut, Laurel Orr, Isabel Papadimitriou,
  Joon~Sung Park, Chris Piech, Eva Portelance, Christopher Potts, Aditi
  Raghunathan, Robert Reich, Hongyu Ren, Frieda Rong, Yusuf~H. Roohani, Camilo
  Ruiz, Jack Ryan, Christopher R'e, Dorsa Sadigh, Shiori Sagawa, Keshav
  Santhanam, Andy Shih, Krishna~Parasuram Srinivasan, Alex Tamkin, Rohan Taori,
  Armin~W. Thomas, Florian Tram{\`e}r, Rose~E. Wang, William Wang, Bohan Wu,
  Jiajun Wu, Yuhuai Wu, Sang~Michael Xie, Michihiro Yasunaga, Jiaxuan You,
  Matei~A. Zaharia, Michael Zhang, Tianyi Zhang, Xikun Zhang, Yuhui Zhang,
  Lucia Zheng, Kaitlyn Zhou, and Percy Liang.
\newblock On the opportunities and risks of foundation models.
\newblock {\em arXiv}, 2021.

\bibitem{bordt2022posthoc}
Sebastian Bordt, Mich{\`{e}}le Finck, Eric Raidl, and Ulrike von Luxburg.
\newblock Post-hoc explanations fail to achieve their purpose in adversarial
  contexts.
\newblock {\em ACM Conference on Fairness, Accountability, and Transparency},
  5, 2022.

\bibitem{pmlr-v206-bordt23a}
Sebastian Bordt and Ulrike von Luxburg.
\newblock From shapley values to generalized additive models and back.
\newblock In Francisco Ruiz, Jennifer Dy, and Jan-Willem van~de Meent, editors,
  {\em Proceedings of The 26th International Conference on Artificial
  Intelligence and Statistics}, volume 206 of {\em Proceedings of Machine
  Learning Research}, pages 709--745. PMLR, 25--27 Apr 2023.

\bibitem{brandt2023precise}
Rafa{\"e}l Brandt, Daan Raatjens, and Georgi Gaydadjiev.
\newblock Precise benchmarking of explainable {AI} attribution methods.
\newblock {\em arXiv preprint arXiv:2308.03161}, 2023.

\bibitem{buolamwiniGenderShadesIntersectional2018}
Joy Buolamwini and Timnit Gebru.
\newblock Gender {{Shades}}: {{Intersectional Accuracy Disparities}} in
  {{Commercial Gender Classification}}.
\newblock In {\em Conference on {{Fairness}}, {{Accountability}} and
  {{Transparency}}}, pages 77--91. {PMLR}, January 2018.

\bibitem{camburuCanTrustExplainer2019}
Oana-Maria Camburu, Eleonora Giunchiglia, Jakob Foerster, Thomas Lukasiewicz,
  and Phil Blunsom.
\newblock Can {{I Trust}} the {{Explainer}}? {{Verifying Post}}-hoc
  {{Explanatory Methods}}.
\newblock {\em NeurIPS 2019 Workshop on Safety and Robustness in Decision
  Making}, 1:1--13, December 2019.

\bibitem{carmichaelposthoceval2021}
Zachariah Carmichael and Walter~J. Scheirer.
\newblock A framework for evaluating post hoc feature-additive explainers.
\newblock {\em arXiv}, abs/2106.08376, 2021.

\bibitem{unfooling2023}
Zachariah Carmichael and Walter~J Scheirer.
\newblock Unfooling perturbation-based post hoc explainers.
\newblock In {\em Proceedings of the {AAAI} Conference on Artificial
  Intelligence}. {AAAI}, 2023.

\bibitem{carter2022explainable}
Selina Carter and Jonathan Hersh.
\newblock Explainable {AI} helps bridge the {AI} skills gap: Evidence from a
  large bank.
\newblock {\em Economics Faculty Articles and Research}, 276, 2022.

\bibitem{doi:10.1148/radiol.230574}
Julius Chapiro.
\newblock Explainable {AI} for prostate {MRI}: Don’t trust, verify.
\newblock {\em Radiology}, 307(4):e230574, 2023.

\bibitem{chaszczewicz2023task}
Alicja Chaszczewicz.
\newblock Is task-agnostic explainable {AI} a myth?
\newblock {\em arXiv preprint arXiv:2307.06963}, 2023.

\bibitem{pmlr-v80-chen18j}
Jianbo Chen, Le~Song, Martin Wainwright, and Michael Jordan.
\newblock Learning to explain: An information-theoretic perspective on model
  interpretation.
\newblock In Jennifer Dy and Andreas Krause, editors, {\em Proceedings of the
  35th International Conference on Machine Learning}, volume~80 of {\em
  Proceedings of Machine Learning Research}, pages 883--892. PMLR, 10--15 Jul
  2018.

\bibitem{NEURIPS2020_ce758408}
Jonathan Crabbe, Yao Zhang, William Zame, and Mihaela van~der Schaar.
\newblock Learning outside the black-box: The pursuit of interpretable models.
\newblock In H.~Larochelle, M.~Ranzato, R.~Hadsell, M.F. Balcan, and H.~Lin,
  editors, {\em Advances in Neural Information Processing Systems}, volume~33,
  pages 17838--17849. Curran Associates, Inc., 2020.

\bibitem{SchumerSAFE}
CSIS.
\newblock {S}en. {C}huck {S}chumer launches {SAFE} innovation in the {AI} age
  at {CSIS}.
\newblock
  \url{https://www.csis.org/analysis/sen-chuck-schumer-launches-safe-innovation-ai-age-csis},
  2023.
\newblock Accessed: 2023-09-13.

\bibitem{DBLP:conf/aaai/BroeckLSS21}
Guy~Van den Broeck, Anton Lykov, Maximilian Schleich, and Dan Suciu.
\newblock On the tractability of {SHAP} explanations.
\newblock In {\em Thirty-Fifth {AAAI} Conference on Artificial Intelligence,
  {AAAI} 2021, Thirty-Third Conference on Innovative Applications of Artificial
  Intelligence, {IAAI} 2021, The Eleventh Symposium on Educational Advances in
  Artificial Intelligence, {EAAI} 2021, Virtual Event, February 2-9, 2021},
  pages 6505--6513. {AAAI} Press, 2021.

\bibitem{deyoungERASER2020}
Jay DeYoung, Sarthak Jain, Nazneen~Fatema Rajani, Eric Lehman, Caiming Xiong,
  Richard Socher, and Byron~C. Wallace.
\newblock {ERASER}: A benchmark to evaluate rationalized {NLP} models.
\newblock {\em Transactions of the Association for Computational Linguistics},
  pages 1--16, July 2020.

\bibitem{dimanov2020you}
Botty Dimanov, Umang Bhatt, Mateja Jamnik, and Adrian Weller.
\newblock You shouldn’t trust me: Learning models which conceal unfairness
  from multiple explanation methods.
\newblock {\em Frontiers in Artificial Intelligence and Applications: {ECAI}},
  2020.

\bibitem{dombrowski2019explanations}
Ann-Kathrin Dombrowski, Maximillian Alber, Christopher Anders, Marcel
  Ackermann, Klaus-Robert M{\"u}ller, and Pan Kessel.
\newblock Explanations can be manipulated and geometry is to blame.
\newblock {\em Advances in neural information processing systems}, 32, 2019.

\bibitem{doshi2017towards}
Finale Doshi-Velez and Been Kim.
\newblock Towards a rigorous science of interpretable machine learning.
\newblock {\em arXiv preprint arXiv:1702.08608}, 2017.

\bibitem{DBLP:conf/acl/DuDX0022}
Li~Du, Xiao Ding, Kai Xiong, Ting Liu, and Bing Qin.
\newblock e-{CARE}: a new dataset for exploring explainable causal reasoning.
\newblock In Smaranda Muresan, Preslav Nakov, and Aline Villavicencio, editors,
  {\em Proceedings of the 60th Annual Meeting of the Association for
  Computational Linguistics (Volume 1: Long Papers), {ACL} 2022, Dublin,
  Ireland, May 22-27, 2022}, pages 432--446. Association for Computational
  Linguistics, 2022.

\bibitem{EU-GDPR}
Council of~the {EU} and European Parliament.
\newblock {R}egulation ({EU}) 2016/679 of the {E}uropean {P}arliament and of
  the {C}ouncil of 27 {A}pril 2016 on the protection of natural persons with
  regard to the processing of personal data and on the free movement of such
  data, and repealing {D}irective 95/46/{EC} ({G}eneral {D}ata {P}rotection
  {R}egulation).
\newblock {\em Official Journal of the European Union}, L 119:1--88, 2016.

\bibitem{EU-AI-Act}
{E}uropean {C}ommission.
\newblock {P}roposal for a regulation of the {E}uropean {P}arliament and the
  {C}ouncil: {L}aying down harmonised rules on {Artificial Intelligence
  (Artificial Intelligence Act)} and amending certain {U}nion legislative acts.
\newblock
  \url{https://eur-lex.europa.eu/legal-content/EN/TXT/?uri=CELEX:52021PC0206},
  4 2021.

\bibitem{10.1145/3533767.3534225}
Ming Fan, Jiali Wei, Wuxia Jin, Zhou Xu, Wenying Wei, and Ting Liu.
\newblock One step further: Evaluating interpreters using metamorphic testing.
\newblock In {\em Proceedings of the 31st ACM SIGSOFT International Symposium
  on Software Testing and Analysis}, ISSTA 2022, page 327–339, New York, NY,
  USA, 2022. Association for Computing Machinery.

\bibitem{fengWhatCanAI2019}
Shi Feng and Jordan {Boyd-Graber}.
\newblock What can {{AI}} do for me? {E}valuating machine learning
  interpretations in cooperative play.
\newblock In Wai{-}Tat Fu, Shimei Pan, Oliver Brdiczka, Polo Chau, and Gaelle
  Calvary, editors, {\em Annual Conference on Intelligent User Interfaces},
  pages 229--239. {ACM}, March 2019.

\bibitem{ficoheloc2018}
Fair Isaac~Corporation (FICO).
\newblock {FICO} explainable machine learning challenge: Home equity line of
  credit ({HELOC}) dataset, 2018.
\newblock
  \url{https://community.fico.com/s/explainable-machine-learning-challenge}.

\bibitem{Friedman2001}
Jerome~H. Friedman.
\newblock Greedy function approximation: A gradient boosting machine.
\newblock {\em The Annals of Statistics}, 29(5):1189--1232, 2001.

\bibitem{NEURIPS2020_0d770c49}
Christopher Frye, Colin Rowat, and Ilya Feige.
\newblock Asymmetric shapley values: incorporating causal knowledge into
  model-agnostic explainability.
\newblock In H.~Larochelle, M.~Ranzato, R.~Hadsell, M.F. Balcan, and H.~Lin,
  editors, {\em Advances in Neural Information Processing Systems}, volume~33,
  pages 1229--1239. Curran Associates, Inc., 2020.

\bibitem{pmlr-v108-garreau20a}
Damien Garreau and Ulrike von Luxburg.
\newblock Explaining the explainer: A first theoretical analysis of {LIME}.
\newblock In Silvia Chiappa and Roberto Calandra, editors, {\em International
  Conference on Artificial Intelligence and Statistics}, volume 108 of {\em
  Proceedings of Machine Learning Research}, pages 1287--1296. {PMLR}, August
  2020.

\bibitem{rpy2}
L~Gautier.
\newblock rpy2: A simple and efficient access to {R} from {P}ython.
\newblock {\em URL http://rpy.sourceforge.net/rpy2.html}, 3:1, 2008.

\bibitem{ge2021counterfactual}
Yingqiang Ge, Shuchang Liu, Zelong Li, Shuyuan Xu, Shijie Geng, Yunqi Li,
  Juntao Tan, Fei Sun, and Yongfeng Zhang.
\newblock Counterfactual evaluation for explainable {AI}.
\newblock {\em arXiv preprint arXiv:2109.01962}, 2021.

\bibitem{geirhos2023don}
Robert Geirhos, Roland~S Zimmermann, Blair Bilodeau, Wieland Brendel, and Been
  Kim.
\newblock Don't trust your eyes: on the (un) reliability of feature
  visualizations.
\newblock {\em arXiv preprint arXiv:2306.04719}, 2023.

\bibitem{guidottiEvaluatingLocalExplanation2021}
Riccardo Guidotti.
\newblock Evaluating local explanation methods on ground truth.
\newblock {\em Artificial Intelligence}, 291:1--16, February 2021.

\bibitem{guidottiSurveyMethodsExplaining2018}
Riccardo Guidotti, Anna Monreale, Salvatore Ruggieri, Franco Turini, Fosca
  Giannotti, and Dino Pedreschi.
\newblock A survey of methods for explaining black box models.
\newblock {\em ACM Computing Surveys}, 51(5):93:1--93:42, August 2018.

\bibitem{2020NumPy-Array}
Charles~R. Harris, K.~Jarrod Millman, Stéfan~J van~der Walt, Ralf Gommers,
  Pauli Virtanen, David Cournapeau, Eric Wieser, Julian Taylor, Sebastian Berg,
  Nathaniel~J. Smith, Robert Kern, Matti Picus, Stephan Hoyer, Marten~H. van
  Kerkwijk, Matthew Brett, Allan Haldane, Jaime Fernández~del Río, Mark
  Wiebe, Pearu Peterson, Pierre Gérard-Marchant, Kevin Sheppard, Tyler Reddy,
  Warren Weckesser, Hameer Abbasi, Christoph Gohlke, and Travis~E. Oliphant.
\newblock Array programming with {NumPy}.
\newblock {\em Nature}, 585:357–362, 2020.

\bibitem{HARRISON197881}
David Harrison, Jr. and Daniel~L Rubinfeld.
\newblock Hedonic housing prices and the demand for clean air.
\newblock {\em Journal of Environmental Economics and Management},
  5(1):81--102, March 1978.

\bibitem{DBLP:books/lib/HastieTF09}
Trevor Hastie, Robert Tibshirani, and Jerome~H. Friedman.
\newblock {\em The Elements of Statistical Learning: Data Mining, Inference,
  and Prediction, 2nd Edition}.
\newblock Springer Series in Statistics. Springer, February 2009.

\bibitem{hatherley2022virtues}
Joshua Hatherley, Robert Sparrow, and Mark Howard.
\newblock The virtues of interpretable medical artificial intelligence.
\newblock {\em Cambridge Quarterly of Healthcare Ethics}, pages 1--10, 2022.

\bibitem{hedstrom2023meta}
Anna Hedstr{\"o}m, Philine Bommer, Kristoffer~K Wickstr{\o}m, Wojciech Samek,
  Sebastian Lapuschkin, and Marina M-C H{\"o}hne.
\newblock The meta-evaluation problem in explainable {AI}: Identifying reliable
  estimators with {M}eta{Q}uantus.
\newblock {\em arXiv preprint arXiv:2302.07265}, 2023.

\bibitem{hendrycksNaturalAdversarialExamples2019}
Dan Hendrycks, Kevin Zhao, Steven Basart, Jacob Steinhardt, and Dawn Song.
\newblock Natural adversarial examples.
\newblock In {\em Conference on Computer Vision and Pattern Recognition}, pages
  15262--15271. {IEEE}/Computer Vision Foundation, 2021.

\bibitem{NEURIPS2019_6950aa02}
Lukas Hoyer, Mauricio Munoz, Prateek Katiyar, Anna Khoreva, and Volker Fischer.
\newblock Grid saliency for context explanations of semantic segmentation.
\newblock In H.~Wallach, H.~Larochelle, A.~Beygelzimer, F.~d\textquotesingle
  Alch\'{e}-Buc, E.~Fox, and R.~Garnett, editors, {\em Advances in Neural
  Information Processing Systems}, volume~32. Curran Associates, Inc., 2019.

\bibitem{hruska2022ground}
Eugen Hruska, Liang Zhao, and Fang Liu.
\newblock Ground truth explanation dataset for chemical property prediction on
  molecular graphs.
\newblock {\em ChemRxiv}, 2022.

\bibitem{humer2022comparing}
Christina Humer, Andreas Hinterreiter, Benedikt Leichtmann, Martina Mara, and
  Marc Streit.
\newblock Comparing effects of attribution-based, example-based, and
  feature-based explanation methods on ai-assisted decision-making.
\newblock {\em OSF Preprints}, 2022.

\bibitem{matplotlib}
John~D. Hunter.
\newblock Matplotlib: A 2d graphics environment.
\newblock {\em Computing in Science Engineering}, 9(3):90--95, 2007.

\bibitem{NEURIPS2020_47a3893c}
Aya~Abdelsalam Ismail, Mohamed Gunady, Hector Corrada~Bravo, and Soheil Feizi.
\newblock Benchmarking deep learning interpretability in time series
  predictions.
\newblock In H.~Larochelle, M.~Ranzato, R.~Hadsell, M.F. Balcan, and H.~Lin,
  editors, {\em Advances in Neural Information Processing Systems}, volume~33,
  pages 6441--6452. Curran Associates, Inc., 2020.

\bibitem{Jain2022Importance}
Vaishali Jain, Ted Enamorado, and Cynthia Rudin.
\newblock The {Importance} of {Being} {Ernest}, {Ekundayo}, or {Eswari}: An
  {Interpretable} {Machine} {Learning} {Approach} to {Name}-{Based} {Ethnicity}
  {Classification}.
\newblock {\em Harvard Data Science Review}, 4(3), jul 28 2022.
\newblock https://hdsr.mitpress.mit.edu/pub/wgss79vu.

\bibitem{jia2019improving}
Yunzhe Jia, James Bailey, Kotagiri Ramamohanarao, Christopher Leckie, and
  Michael~E Houle.
\newblock Improving the quality of explanations with local embedding
  perturbations.
\newblock In {\em Proceedings of the 25th ACM SIGKDD International conference
  on knowledge discovery \& Data Mining}, pages 875--884, 2019.

\bibitem{Jia2020}
Yunzhe Jia, James Bailey, Kotagiri Ramamohanarao, Christopher Leckie, and
  Xingjun Ma.
\newblock Exploiting patterns to explain individual predictions.
\newblock {\em Knowledge and Information Systems}, 62(3):927--950, Mar 2020.

\bibitem{PDPbox}
Li~Jiangchun, Carlos Daniel~C Santos, Michael Kuhlen, and Angertdev Singh.
\newblock Pdpbox: v0.2.1, March 2021.

\bibitem{jin2023rethinking}
Weina Jin, Xiaoxiao Li, and Ghassan Hamarneh.
\newblock The {XAI} alignment problem: Rethinking how should we evaluate
  human-centered {AI} explainability techniques.
\newblock {\em arXiv preprint arXiv:2303.17707}, 2023.

\bibitem{Jin2020Towards}
Xisen Jin, Zhongyu Wei, Junyi Du, Xiangyang Xue, and Xiang Ren.
\newblock Towards hierarchical importance attribution: Explaining compositional
  semantics for neural sequence models.
\newblock In {\em International Conference on Learning Representations}, 2020.

\bibitem{joblib}
{Joblib Development Team}.
\newblock Joblib: running python functions as pipeline jobs, 2020.

\bibitem{mpmath}
Fredrik Johansson et~al.
\newblock {\em mpmath: a {P}ython library for arbitrary-precision
  floating-point arithmetic (version 0.18)}, December 2013.
\newblock {\tt http://mpmath.org/}.

\bibitem{kaurInterpretingInterpretabilityUnderstanding2020}
Harmanpreet Kaur, Harsha Nori, Samuel Jenkins, Rich Caruana, Hanna Wallach, and
  Jennifer Wortman~Vaughan.
\newblock Interpreting interpretability: Understanding data scientists' use of
  interpretability tools for machine learning.
\newblock In Regina Bernhaupt, Florian~'Floyd' Mueller, David Verweij, Josh
  Andres, Joanna McGrenere, Andy Cockburn, Ignacio Avellino, Alix Goguey,
  Pernille Bj{\o}n, Shengdong Zhao, Briane~Paul Samson, and Rafal Kocielnik,
  editors, {\em Conference on Human Factors in Computing Systems}, pages 1--14.
  {ACM}, April 2020.

\bibitem{pmlr-v80-kim18d}
Been Kim, Martin Wattenberg, Justin Gilmer, Carrie Cai, James Wexler, Fernanda
  Viegas, and Rory sayres.
\newblock Interpretability beyond feature attribution: Quantitative testing
  with concept activation vectors ({TCAV}).
\newblock In Jennifer Dy and Andreas Krause, editors, {\em Proceedings of the
  35th International Conference on Machine Learning}, volume~80 of {\em
  Proceedings of Machine Learning Research}, pages 2668--2677. PMLR, 10--15 Jul
  2018.

\bibitem{alibi}
Janis Klaise, Arnaud Van~Looveren, Giovanni Vacanti, and Alexandru Coca.
\newblock Alibi: Algorithms for monitoring and explaining machine learning
  models, 2019.

\bibitem{krishna2022disagreement}
Satyapriya Krishna, Tessa Han, Alex Gu, Javin Pombra, Shahin Jabbari, Steven
  Wu, and Himabindu Lakkaraju.
\newblock The disagreement problem in explainable machine learning: A
  practitioner's perspective.
\newblock {\em arXiv:2202.01602}, pages 1--46, 2022.

\bibitem{pmlr-v119-lakkaraju20a}
Himabindu Lakkaraju, Nino Arsov, and Osbert Bastani.
\newblock Robust and stable black box explanations.
\newblock In Hal~Daumé III and Aarti Singh, editors, {\em Proceedings of the
  37th International Conference on Machine Learning}, volume 119 of {\em
  Proceedings of Machine Learning Research}, pages 5628--5638. PMLR, 13--18 Jul
  2020.

\bibitem{Lapuschkin_2016_CVPR}
Sebastian Lapuschkin, Alexander Binder, Gregoire Montavon, Klaus-Robert Muller,
  and Wojciech Samek.
\newblock Analyzing classifiers: {F}isher vectors and deep neural networks.
\newblock In {\em Proceedings of the IEEE Conference on Computer Vision and
  Pattern Recognition (CVPR)}, June 2016.

\bibitem{leavitt2020towards}
Matthew~L. Leavitt and Ari Morcos.
\newblock Towards falsifiable interpretability research.
\newblock In {\em {NeurIPS} Workshop on {ML}-Retrospectives, Surveys \&
  Meta-Analyses}, pages 1--15. arXiv, 2020.

\bibitem{MNIST}
Yann LeCun, Léon Bottou, Yoshua Bengio, and Patrick Haffner.
\newblock Gradient-based learning applied to document recognition.
\newblock {\em Proceedings of the IEEE}, 86(11):2278--2324, 1998.

\bibitem{US-Alg-Acct-Act}
{L}ibrary~of {C}ongress.
\newblock {H.R.6580} - 117th {C}ongress (2021-2022): Algorithmic accountability
  act of 2022.
\newblock
  \url{https://www.congress.gov/bill/117th-congress/house-bill/6580/text}, 2
  2022.

\bibitem{liptonMythos2018}
Zachary~C. Lipton.
\newblock The mythos of model interpretability.
\newblock {\em {ACM} Queue}, 16(3):30, July 2018.

\bibitem{10.5555/3304889.3305002}
Ninghao Liu, Donghwa Shin, and Xia Hu.
\newblock Contextual outlier interpretation.
\newblock In {\em Proceedings of the 27th International Joint Conference on
  Artificial Intelligence}, IJCAI'18, page 2461–2467. AAAI Press, 2018.

\bibitem{lundbergUnifiedApproachInterpreting2017}
Scott~M. Lundberg and Su-In Lee.
\newblock A unified approach to interpreting model predictions.
\newblock In Isabelle Guyon, Ulrike von Luxburg, Samy Bengio, Hanna~M. Wallach,
  Rob Fergus, S.~V.~N. Vishwanathan, and Roman Garnett, editors, {\em
  Conference on Neural Information Processing Systems}, pages 4765--4774,
  December 2017.

\bibitem{NEURIPS2020_e37b08dd}
Dongsheng Luo, Wei Cheng, Dongkuan Xu, Wenchao Yu, Bo~Zong, Haifeng Chen, and
  Xiang Zhang.
\newblock Parameterized explainer for graph neural network.
\newblock In H.~Larochelle, M.~Ranzato, R.~Hadsell, M.F. Balcan, and H.~Lin,
  editors, {\em Advances in Neural Information Processing Systems}, volume~33,
  pages 19620--19631. Curran Associates, Inc., 2020.

\bibitem{incidentDB}
Sean McGregor.
\newblock Preventing repeated real world {AI} failures by cataloging incidents:
  The {AI} incident database.
\newblock In {\em {AAAI} Conference on Innovative Applications of Artificial
  Intelligence, {IAAI}}, pages 15458--15463. {AAAI} Press, 2021.

\bibitem{sympy}
Aaron Meurer, Christopher~P. Smith, Mateusz Paprocki, Ond\v{r}ej
  \v{C}ert\'{i}k, Sergey~B. Kirpichev, Matthew Rocklin, Amit Kumar, Sergiu
  Ivanov, Jason~K. Moore, Sartaj Singh, Thilina Rathnayake, Sean Vig, Brian~E.
  Granger, Richard~P. Muller, Francesco Bonazzi, Harsh Gupta, Shivam Vats,
  Fredrik Johansson, Fabian Pedregosa, Matthew~J. Curry, Andy~R. Terrel,
  \v{S}t\v{e}p\'{a}n Rou\v{c}ka, Ashutosh Saboo, Isuru Fernando, Sumith Kulal,
  Robert Cimrman, and Anthony Scopatz.
\newblock {S}ym{P}y: symbolic computing in {P}ython.
\newblock {\em PeerJ Computer Science}, 3:e103, January 2017.

\bibitem{millerAIMedical2018}
D.~Douglas Miller and Eric~W. Brown.
\newblock Artificial intelligence in medical practice: The question to the
  answer?
\newblock {\em The American Journal of Medicine}, 131(2):129--133, 2018.

\bibitem{miro2023novel}
Miquel Mir{\'o}-Nicolau, Antoni Jaume-i Cap{\'o}, and Gabriel Moy{\`a}-Alcover.
\newblock A novel approach to generate datasets with {XAI} ground truth to
  evaluate image models.
\newblock {\em arXiv preprint arXiv:2302.05624}, 2023.

\bibitem{molnar2019}
Christoph Molnar.
\newblock {\em Interpretable Machine Learning}.
\newblock Leanpub, 2019.
\newblock \url{https://christophm.github.io/interpretable-ml-book/}.

\bibitem{DBLP:conf/lwa/MuckeP22}
Sascha M{\"{u}}cke and Lukas Pfahler.
\newblock Check mate: {A} sanity check for trustworthy {AI}.
\newblock In Pascal Reuss, Viktor Eisenstadt, Jakob~Michael Sch{\"{o}}nborn,
  and Jero Sch{\"{a}}fer, editors, {\em Proceedings of the {LWDA} 2022
  Workshops: FGWM, FGKD, and FGDB, Hildesheim (Germany), Oktober 5-7th, 2022},
  volume 3341 of {\em {CEUR} Workshop Proceedings}, pages 91--103. CEUR-WS.org,
  2022.

\bibitem{nam2020relative}
Woo-Jeoung Nam, Shir Gur, Jaesik Choi, Lior Wolf, and Seong-Whan Lee.
\newblock Relative attributing propagation: Interpreting the comparative
  contributions of individual units in deep neural networks.
\newblock In {\em Proceedings of the AAAI conference on artificial
  intelligence}, volume~34, pages 2501--2508, 2020.

\bibitem{NEURIPS2020_426f990b}
Karthikeyan Natesan~Ramamurthy, Bhanukiran Vinzamuri, Yunfeng Zhang, and Amit
  Dhurandhar.
\newblock Model agnostic multilevel explanations.
\newblock In H.~Larochelle, M.~Ranzato, R.~Hadsell, M.F. Balcan, and H.~Lin,
  editors, {\em Advances in Neural Information Processing Systems}, volume~33,
  pages 5968--5979. Curran Associates, Inc., 2020.

\bibitem{anecdotalEvidenceXAI2023}
Meike Nauta, Jan Trienes, Shreyasi Pathak, Elisa Nguyen, Michelle Peters,
  Yasmin Schmitt, J\"{o}rg Schl\"{o}tterer, Maurice van Keulen, and Christin
  Seifert.
\newblock From anecdotal evidence to quantitative evaluation methods: A
  systematic review on evaluating explainable {AI}.
\newblock {\em ACM Comput. Surv.}, 55(13s), jul 2023.

\bibitem{oneilWeaponsMathDestruction2016}
Cathy O'Neil.
\newblock {\em Weapons of {{Math Destruction}}: {{How Big Data Increases
  Inequality}} and {{Threatens Democracy}}}.
\newblock {Crown}, September 2016.

\bibitem{oramas2018visual}
Jose Oramas, Kaili Wang, and Tinne Tuytelaars.
\newblock Visual explanation by interpretation: Improving visual feedback
  capabilities of deep neural networks.
\newblock In {\em International Conference on Learning Representations}, 2019.

\bibitem{pawelczyk2021carla}
Martin Pawelczyk, Sascha Bielawski, Johannes van~den Heuvel, Tobias Richter,
  and Gjergji Kasneci.
\newblock Carla: a {P}ython library to benchmark algorithmic recourse and
  counterfactual explanation algorithms.
\newblock {\em arXiv:2108.00783}, pages 1--22, 2021.

\bibitem{sklearn}
Fabian Pedregosa, Ga{{\"e}}l Varoquaux, Alexandre Gramfort, Vincent Michel,
  Bertrand Thirion, Olivier Grisel, Mathieu Blondel, Peter Prettenhofer, Ron
  Weiss, Vincent Dubourg, Jake Vanderplas, Alexandre Passos, David Cournapeau,
  Matthieu Brucher, Matthieu Perrot, and {{\'E}}douard Duchesnay.
\newblock Scikit-learn: Machine learning in python.
\newblock {\em Journal of Machine Learning Research}, 12(85):2825--2830, 2011.

\bibitem{peters2023unjustified}
Uwe Peters and Mary Carman.
\newblock Unjustified sample sizes and generalizations in explainable {AI}
  research: Principles for more inclusive user studies.
\newblock {\em arXiv preprint arXiv:2305.09477}, 2023.

\bibitem{plumbModelAgnosticSupervised2018}
Gregory Plumb, Denali Molitor, and Ameet~S. Talwalkar.
\newblock Model agnostic supervised local explanations.
\newblock In Samy Bengio, Hanna~M. Wallach, Hugo Larochelle, Kristen Grauman,
  Nicol{\`{o}} Cesa{-}Bianchi, and Roman Garnett, editors, {\em Conference on
  Neural Information Processing Systems}, pages 2520--2529, December 2018.

\bibitem{pmlr-v119-plumb20a}
Gregory Plumb, Jonathan Terhorst, Sriram Sankararaman, and Ameet Talwalkar.
\newblock Explaining groups of points in low-dimensional representations.
\newblock In Hal~Daumé III and Aarti Singh, editors, {\em Proceedings of the
  37th International Conference on Machine Learning}, volume 119 of {\em
  Proceedings of Machine Learning Research}, pages 7762--7771. PMLR, 13--18 Jul
  2020.

\bibitem{poerner-etal-2018-evaluating}
Nina Poerner, Hinrich Sch{\"u}tze, and Benjamin Roth.
\newblock Evaluating neural network explanation methods using hybrid documents
  and morphosyntactic agreement.
\newblock In {\em Proceedings of the 56th Annual Meeting of the Association for
  Computational Linguistics (Volume 1: Long Papers)}, pages 340--350,
  Melbourne, Australia, July 2018. Association for Computational Linguistics.

\bibitem{pruthi-etal-2020-learning}
Danish Pruthi, Mansi Gupta, Bhuwan Dhingra, Graham Neubig, and Zachary~C.
  Lipton.
\newblock Learning to deceive with attention-based explanations.
\newblock In {\em Proceedings of the 58th Annual Meeting of the Association for
  Computational Linguistics}, pages 4782--4793, Online, July 2020. Association
  for Computational Linguistics.

\bibitem{quionero-candelaDatasetShiftMachine2009}
Joaquin {Quionero-Candela}, Masashi Sugiyama, Anton Schwaighofer, and Neil~D.
  Lawrence.
\newblock {\em Dataset {{Shift}} in {{Machine Learning}}}.
\newblock {The MIT Press}, 2009.

\bibitem{R}
{R Core Team}.
\newblock {\em R: A Language and Environment for Statistical Computing}.
\newblock R Foundation for Statistical Computing, Vienna, Austria, 2021.

\bibitem{DBLP:journals/adac/RamonMPE20}
Yanou Ramon, David Martens, Foster~J. Provost, and Theodoros Evgeniou.
\newblock A comparison of instance-level counterfactual explanation algorithms
  for behavioral and textual data: Sedc, {LIME-C} and {SHAP-C}.
\newblock {\em Advances in Data Analysis and Classification}, 14(4):801--819,
  2020.

\bibitem{rasmussen2023machines}
Maria~H Rasmussen, Diana~S Christensen, and Jan~H Jensen.
\newblock Do machines dream of atoms? {C}rippen's log{P} as a quantitative
  molecular benchmark for explainable {AI} heatmaps.
\newblock {\em SciPost Chemistry}, 2(1):002, 2023.

\bibitem{ribeiroWhyShouldTrust2016}
Marco~Tulio Ribeiro, Sameer Singh, and Carlos Guestrin.
\newblock ``{Why} should {I} trust you?'': Explaining the predictions of any
  classifier.
\newblock In {\em SIGKDD International Conference on Knowledge Discovery \&
  Data Mining}, pages 1135--1144. {ACM}, August 2016.

\bibitem{10.5555/3172077.3172259}
Andrew~Slavin Ross, Michael~C. Hughes, and Finale Doshi-Velez.
\newblock Right for the right reasons: Training differentiable models by
  constraining their explanations.
\newblock In {\em Proceedings of the 26th International Joint Conference on
  Artificial Intelligence}, IJCAI'17, page 2662–2670. AAAI Press, 2017.

\bibitem{Rudin2019}
Cynthia Rudin.
\newblock Stop explaining black box machine learning models for high stakes
  decisions and use interpretable models instead.
\newblock {\em Nature Machine Intelligence}, 1(5):206--215, May 2019.

\bibitem{rudin2022black}
Cynthia Rudin.
\newblock Why black box machine learning should be avoided for high-stakes
  decisions, in brief.
\newblock {\em Nature Reviews Methods Primers}, 2(1):81, 2022.

\bibitem{Saporta2022}
Adriel Saporta, Xiaotong Gui, Ashwin Agrawal, Anuj Pareek, Steven Q.~H. Truong,
  Chanh D.~T. Nguyen, Van-Doan Ngo, Jayne Seekins, Francis~G. Blankenberg,
  Andrew~Y. Ng, Matthew~P. Lungren, and Pranav Rajpurkar.
\newblock Benchmarking saliency methods for chest {X}-ray interpretation.
\newblock {\em Nature Machine Intelligence}, 4(10):867--878, Oct 2022.

\bibitem{Schwalbe2023}
Gesina Schwalbe and Bettina Finzel.
\newblock A comprehensive taxonomy for explainable artificial intelligence: a
  systematic survey of surveys on methods and concepts.
\newblock {\em Data Mining and Knowledge Discovery}, Jan 2023.

\bibitem{shapr}
Nikolai Sellereite, Martin Jullum, and Annabelle Redelmeier.
\newblock {\em shapr: Prediction Explanation with Dependence-Aware Shapley
  Values}, 2021.
\newblock R package version 0.2.0.

\bibitem{selvarajuGradCAMVisualExplanations2017}
R.~R. Selvaraju, M.~Cogswell, A.~Das, R.~Vedantam, D.~Parikh, and D.~Batra.
\newblock Grad-{{CAM}}: {{Visual Explanations}} from {{Deep Networks}} via
  {{Gradient}}-{{Based Localization}}.
\newblock In {\em 2017 {{IEEE International Conference}} on {{Computer Vision}}
  ({{ICCV}})}, pages 618--626, October 2017.

\bibitem{pygam}
Daniel Servén, Charlie Brummitt, Hassan Abedi, and hlink.
\newblock dswah/pygam: v0.8.0, October 2018.

\bibitem{slackFoolingLIMESHAP2020}
Dylan Slack, Sophie Hilgard, Emily Jia, Sameer Singh, and Himabindu Lakkaraju.
\newblock Fooling {{LIME}} and {{SHAP}}: Adversarial attacks on post hoc
  explanation methods.
\newblock In Annette~N. Markham, Julia Powles, Toby Walsh, and Anne~L.
  Washington, editors, {\em Conference on AI, Ethics, and Society}, pages
  180--186. AAAI/ACM, February 2020.

\bibitem{Subramanya_2019_ICCV}
Akshayvarun Subramanya, Vipin Pillai, and Hamed Pirsiavash.
\newblock Fooling network interpretation in image classification.
\newblock In {\em Proceedings of the IEEE/CVF International Conference on
  Computer Vision (ICCV)}, October 2019.

\bibitem{swamy2023future}
Vinitra Swamy, Jibril Frej, and Tanja K{\"a}ser.
\newblock The future of human-centric e{X}plainable artificial intelligence
  ({XAI}) is not post-hoc explanations.
\newblock {\em arXiv preprint arXiv:2307.00364}, 2023.

\bibitem{sydorova-etal-2019-interpretable}
Alona Sydorova, Nina Poerner, and Benjamin Roth.
\newblock Interpretable question answering on knowledge bases and text.
\newblock In {\em Proceedings of the 57th Annual Meeting of the Association for
  Computational Linguistics}, pages 4943--4951, Florence, Italy, July 2019.
  Association for Computational Linguistics.

\bibitem{DBLP:journals/corr/SzegedyZSBEGF13}
Christian Szegedy, Wojciech Zaremba, Ilya Sutskever, Joan Bruna, Dumitru Erhan,
  Ian~J. Goodfellow, and Rob Fergus.
\newblock Intriguing properties of neural networks.
\newblock In Yoshua Bengio and Yann LeCun, editors, {\em International
  Conference on Learning Representations}, pages 1--10, April 2014.

\bibitem{adversarialExamples}
Christian Szegedy, Wojciech Zaremba, Ilya Sutskever, Joan Bruna, Dumitru Erhan,
  Ian~J. Goodfellow, and Rob Fergus.
\newblock Intriguing properties of neural networks.
\newblock In Yoshua Bengio and Yann LeCun, editors, {\em International
  Conference on Learning Representations}, pages 1--10, April 2014.

\bibitem{medicalXAIsurvey2020}
Erico Tjoa and Cuntai Guan.
\newblock A survey on explainable artificial intelligence {(XAI):} toward
  medical {XAI}.
\newblock {\em {IEEE} Transactions on Neural Networks and Learning Systems},
  32(11):4793--4813, 2020.

\bibitem{NEURIPS2020_443dec30}
Michael Tsang, Sirisha Rambhatla, and Yan Liu.
\newblock How does this interaction affect me? interpretable attribution for
  feature interactions.
\newblock In H.~Larochelle, M.~Ranzato, R.~Hadsell, M.F. Balcan, and H.~Lin,
  editors, {\em Advances in Neural Information Processing Systems}, volume~33,
  pages 6147--6159. Curran Associates, Inc., 2020.

\bibitem{EU-US-TTC_statement2021}
{U.S.-EU TTC}.
\newblock {U.S.-EU} joint statement of the {T}rade and {T}echnology {C}ouncil.
\newblock
  \url{https://www.commerce.gov/news/press-releases/2022/05/us-eu-joint-statement-trade-and-technology-council},
  5 2022.
\newblock Accessed: 2022-05-10.

\bibitem{van1995python}
Guido Van~Rossum and Fred~L Drake~Jr.
\newblock {\em Python reference manual}.
\newblock Centrum voor Wiskunde en Informatica Amsterdam, 1995.

\bibitem{DBLP:journals/corr/abs-2010-10596}
Sahil Verma, John~P. Dickerson, and Keegan Hines.
\newblock Counterfactual explanations for machine learning: {A} review.
\newblock {\em arXiv:2010.10596}, pages 1--13, 2020.

\bibitem{Vermeire2022}
Tom Vermeire, Dieter Brughmans, Sofie Goethals, Raphael Mazzine~Barbossa
  de~Oliveira, and David Martens.
\newblock Explainable image classification with evidence counterfactual.
\newblock {\em Pattern Analysis and Applications}, 26:1--21, Jan 2022.

\bibitem{2020SciPy-NMeth}
Pauli Virtanen, Ralf Gommers, Travis~E. Oliphant, Matt Haberland, Tyler Reddy,
  David Cournapeau, Evgeni Burovski, Pearu Peterson, Warren Weckesser, Jonathan
  Bright, St{\'e}fan~J. {van der Walt}, Matthew Brett, Joshua Wilson, K.~Jarrod
  Millman, Nikolay Mayorov, Andrew R.~J. Nelson, Eric Jones, Robert Kern, Eric
  Larson, C~J Carey, {\.I}lhan Polat, Yu~Feng, Eric~W. Moore, Jake
  {VanderPlas}, Denis Laxalde, Josef Perktold, Robert Cimrman, Ian Henriksen,
  E.~A. Quintero, Charles~R. Harris, Anne~M. Archibald, Ant{\^o}nio~H. Ribeiro,
  Fabian Pedregosa, Paul {van Mulbregt}, and {SciPy 1.0 Contributors}.
\newblock {{SciPy} 1.0: Fundamental Algorithms for Scientific Computing in
  Python}.
\newblock {\em Nature Methods}, 17:261--272, 2020.

\bibitem{NEURIPS2020_8fb134f2}
Minh Vu and My~T. Thai.
\newblock {PGM-Explainer}: Probabilistic graphical model explanations for graph
  neural networks.
\newblock In H.~Larochelle, M.~Ranzato, R.~Hadsell, M.F. Balcan, and H.~Lin,
  editors, {\em Advances in Neural Information Processing Systems}, volume~33,
  pages 12225--12235. Curran Associates, Inc., 2020.

\bibitem{seaborn}
Michael~L. Waskom.
\newblock seaborn: statistical data visualization.
\newblock {\em Journal of Open Source Software}, 6(60):3021, 2021.

\bibitem{pandas}
{W}es {M}c{K}inney.
\newblock {D}ata {S}tructures for {S}tatistical {C}omputing in {P}ython.
\newblock In {S}t\'efan van~der {W}alt and {J}arrod {M}illman, editors, {\em
  {P}roceedings of the 9th {P}ython in {S}cience {C}onference}, pages 56--61,
  2010.

\bibitem{DBLP:conf/ecai/WhiteG20}
Adam White and Artur~S. d'Avila Garcez.
\newblock Measurable counterfactual local explanations for any classifier.
\newblock In Giuseppe~De Giacomo, Alejandro Catal{\'{a}}, Bistra Dilkina,
  Michela Milano, Sen{\'{e}}n Barro, Alberto Bugar{\'{\i}}n, and
  J{\'{e}}r{\^{o}}me Lang, editors, {\em European Conference on Artificial
  Intelligence}, volume 325 of {\em Frontiers in Artificial Intelligence and
  Applications}, pages 2529--2535. {IOS} Press, 2020.

\bibitem{DBLP:conf/nips/WiegreffeM21}
Sarah Wiegreffe and Ana Marasovic.
\newblock Teach me to explain: {A} review of datasets for explainable natural
  language processing.
\newblock In Joaquin Vanschoren and Sai{-}Kit Yeung, editors, {\em Proceedings
  of the Neural Information Processing Systems Track on Datasets and Benchmarks
  1, NeurIPS Datasets and Benchmarks}, 2021.

\bibitem{yalcin2021evaluating}
Orcun Yalcin, Xiuyi Fan, and Siyuan Liu.
\newblock Evaluating the correctness of explainable {AI} algorithms for
  classification.
\newblock {\em arXiv preprint arXiv:2105.09740}, 2021.

\bibitem{NEURIPS2019_d80b7040}
Zhitao Ying, Dylan Bourgeois, Jiaxuan You, Marinka Zitnik, and Jure Leskovec.
\newblock {GNNExplainer}: Generating explanations for graph neural networks.
\newblock In H.~Wallach, H.~Larochelle, A.~Beygelzimer, F.~d\textquotesingle
  Alch\'{e}-Buc, E.~Fox, and R.~Garnett, editors, {\em Advances in Neural
  Information Processing Systems}, volume~32. Curran Associates, Inc., 2019.

\bibitem{zhangInterpretingCNNsDecision2019}
Quanshi Zhang, Yu~Yang, Haotian Ma, and Ying~Nian Wu.
\newblock Interpreting {{CNNs}} via {{Decision Trees}}.
\newblock In {\em {{Computer Vision}} and {{Pattern Recognition}}}, pages
  6261--6270. IEEE, 2019.

\bibitem{zhang2019should}
Yujia Zhang, Kuangyan Song, Yiming Sun, Sarah Tan, and Madeleine Udell.
\newblock ``why should you trust my explanation?'' understanding uncertainty in
  {LIME} explanations.
\newblock {\em arXiv preprint arXiv:1904.12991}, 2019.

\bibitem{zhouEvaluatingQualityMachine2021}
Jianlong Zhou, Amir~H. Gandomi, Fang Chen, and Andreas Holzinger.
\newblock Evaluating the {{Quality}} of {{Machine Learning Explanations}}: {{A
  Survey}} on {{Methods}} and {{Metrics}}.
\newblock {\em Electronics}, 10(5):593, January 2021.

\bibitem{zhouFeatureAttributionMethods2021}
Yilun Zhou, Serena Booth, Marco~Tulio Ribeiro, and Julie Shah.
\newblock Do {{Feature Attribution Methods Correctly Attribute Features}}?
\newblock {\em {NeurIPS} 1st Workshop on e{X}plainable {AI} approaches for
  debugging and diagnosis ({XAI4Debugging})}, pages 1--22, 2021.

\bibitem{zhou2019ModelAgnostic}
Zihan Zhou, Mingxuan Sun, and Jianhua Chen.
\newblock A model-agnostic approach for explaining the predictions on clustered
  data.
\newblock In {\em 2019 IEEE International Conference on Data Mining (ICDM)},
  pages 1528--1533, 2019.

\end{thebibliography}
}

\clearpage
\appendix
\section*{{\LARGE Supplemental Material}}

\doparttoc %
\faketableofcontents %
\appendix

\part{} %
\parttoc %

\newpage

\section{Expanded Details}

\subsection*{Comprehensive Related Work}
There are three main types of evaluations: application-grounded (real humans, real tasks), human-grounded (real humans, simplified tasks), and functionally-grounded (no humans, proxy tasks)~\cite{doshi2017towards}.
Human- and application-grounded evaluations are expensive, subjective, and qualitative. However, they measure the human utility and effectiveness of explanations. Functionally-grounded metrics are concerned with proxies for the same objectives, but also can quantitatively score the fidelity of an explanation with respect to the model being explained~\cite{anecdotalEvidenceXAI2023}.
The ``Co-12'' explanation properties include aspects about the user (\eg{}, context and controllability), presentation (\eg{}, compactness and confidence), and content (\eg{}, correctness and consistency)~\cite{anecdotalEvidenceXAI2023}. Whereas content-based explanation properties evaluate desirable proxy characteristics, such as the complexity of feature interactions (covariate complexity) or the similarity of explanations between like examples (continuity), evaluations of correctness, or \textit{fidelity}, concern nothing but the explanation faithfulness with respect to the model being explained.
If an explanation is unfaithful to the model, then the question of human utility, trust, or otherwise is irrelevant~\cite{anecdotalEvidenceXAI2023,jin2023rethinking,leavitt2020towards}.
Thus, we are interested in the \textit{functionally-grounded} evaluation of explanation \textit{fidelity} in this paper.

\paragraph{Prior Evaluations of Explanation Fidelity}
While there are evaluation methods for post hoc explainers that act as surrogate predictors (\eg{}, by knowledge distillation) and models that jointly predict and explain~\cite{info14080469,anecdotalEvidenceXAI2023}, we keep our review focused on fidelity evaluations applicable to the explainers detailed in Section~\ref{sec:posthoc-background}.
\textit{Perturbation}-based approaches perturb the model or data and verify that the explanation changes proportionally to either the perturbation or the model~\cite{info14080469,anecdotalEvidenceXAI2023}.
\textit{Removal}-based approaches delete or mask input feature(s) and measure the correlation between model output change and explanation importance score~\cite{info14080469,anecdotalEvidenceXAI2023}. This can be done with single features, or incrementally with feature ordering determined by the explanation importance scores.
The removal can also be accomplished via pixel-flipping, baseline substitutions, zero-padding, or cropping.
While these approaches guarantee that explanations have certain desirable proxy properties, they do not guarantee that explainers are faithful to the exact model behavior~\cite{anecdotalEvidenceXAI2023,jin2023rethinking,leavitt2020towards}. That is unless the model is modified to have those constraints imposed and verified~\cite{bhalla2023verifiable}. However, the question of explanation descriptive completeness of model behavior would still remain~\cite{anecdotalEvidenceXAI2023}.
In turn, we are interested in fidelity evaluations using \textit{ground truth} evaluations. However, this term is overloaded in the literature, so we delineate the three ways it is used here:
\begin{itemize}[noitemsep,nolistsep,leftmargin=*]
\item \textbf{Annotation checks} measure the correlation between feature importance scores and annotated data that is deemed important to the task. Existing evaluations use human annotations, whether it is sample-wise annotations or a crafted annotation-generation process, that evaluate explainers via a proxy task, \eg{}, object localization or rationale generation~\cite{Lapuschkin_2016_CVPR,nam2020relative,Baumgartner_2018_CVPR,NEURIPS2020_56f9f889,NEURIPS2020_443dec30,Saporta2022,poerner-etal-2018-evaluating,Subramanya_2019_ICCV,sydorova-etal-2019-interpretable,10.1145/3533767.3534225,DBLP:conf/acl/DuDX0022,DBLP:conf/lwa/MuckeP22,hruska2022ground,DBLP:conf/nips/WiegreffeM21,camburuCanTrustExplainer2019}. However, this quantifies explanation plausibility rather than fidelity with respect to the model.

\item \textbf{Controlled data checks} involve creating a dataset (typically synthetic) such that a well-performing model should follow some a priori reasoning. For example, this reasoning could be a region of an image belonging to an object of interest, a set of nodes in a graph belonging to a discriminative motif, or a subset of features that are deemed highly discriminative. Two types of evaluations have been studied:
\begin{itemize}[noitemsep,nolistsep,leftmargin=*]
\item \textit{Feature selection} studies evaluate whether an explanation captures a subset of the important features according to the a priori reasoning~\cite{NEURIPS2020_075b051e,NEURIPS2019_6950aa02,pmlr-v80-kim18d,oramas2018visual,NEURIPS2020_0d770c49,10.5555/3304889.3305002,NEURIPS2020_e37b08dd,NEURIPS2019_d80b7040,NEURIPS2020_8fb134f2,pmlr-v119-plumb20a,pruthi-etal-2020-learning,10.5555/3172077.3172259,agarwal2022openxai,rasmussen2023machines,yalcin2021evaluating,zhouFeatureAttributionMethods2021}.
\item \textit{Feature ranking} studies evaluate whether an explanation ranks the importance of (a subset of) features according to the a priori reasoning~\cite{NEURIPS2020_47a3893c,pmlr-v80-chen18j}.
\end{itemize}
However, there is no guarantee that the model actually follows the a priori reasoning as it is still a black-box, even if it performs well on the data.

\item \textbf{White box checks} evaluate the correspondence between explanations and the known white box reasoning. There are several types of evaluations that have
\begin{itemize}[noitemsep,nolistsep,leftmargin=*]
\item \textit{Feature selection} approaches isolate a subset of features that the model uses, \eg{}, by having the model use a subset of features globally, only considering the features leading to the predicted leaf of a tree, or only considering the features that a rule comprises~\cite{brandt2023precise,Jia2020,jia2019improving,zhang2019should}.
\item \textit{Feature ranking} approaches identify the relative importance of each feature so they can be ranked, such as via the coefficients of a logistic regression model~\cite{ge2021counterfactual,ribeiroWhyShouldTrust2016,zhou2019ModelAgnostic}.
\item \textit{Inexact feature contributions} methods use a proxy measure to estimate feature contributions from a model, such as the gradient with respect to each feature in differentiable models or the Gini impurity of trees~\cite{Jin2020Towards,guidottiEvaluatingLocalExplanation2021,NEURIPS2020_426f990b}.
\item \underline{\textit{Exact feature contributions}} methods identify the exact amount each feature contributes to the predicted outcome of a model with respect to a formal (and useful) definition of contribution. Typically, and in this paper, a \textit{contribution} is the amount that a feature (or subset of features) at a particular value adds to the predicted outcome such that the sum of all contributions totals the model prediction.
For instance, the prediction could be the number of times a non-overlapping pattern appears in an image, thus the additive contribution of each pixel is known~\cite{miro2023novel}. Alternatively, the contributions can be taken from a linear regression model~\cite{NEURIPS2020_ce758408,pmlr-v119-lakkaraju20a}.
In~\cite{pmlr-v206-bordt23a}, a connection it is shown that GAMs (with or without interaction effects) can be recovered from Shapley values (with or without interaction effects). They demonstrate that interaction effects with an order of two can be precisely estimated, but can only be detected at higher orders with Shapley values in experiments.
\end{itemize}
White box checks offer the highest fidelity estimate of explanation fidelity as the form of the model, and thus how it uses the data, is well-understood. However, the feature selection and ranking approaches are limited in that the contribution of each feature is unknown. Exact feature contribution approaches are also able to evaluate both feature selection and ranking. In addition, they are more faithful to the true feature contributions within the explained model than inexact feature contributions as no proxy is needed. Our approach fits into this category of white box checks.
\end{itemize}
Our approach has several advantages over prior work.
(1) First, \cite{NEURIPS2020_ce758408,pmlr-v119-lakkaraju20a} define the exact feature contributions as the coefficients of a linear regression model to evaluate the fidelity of explainers, such as \LIME{} and \SHAP{}. However, this is inappropriate for these explainers and thus does not provide an exact set of feature contributions. \LIME{} yields explanations as linear regression coefficients on normalized data, which must be taken into consideration for computing error. \SHAP{} yields explanations as feature-additive contributions rather than coefficients -- for a linear model, the corrected feature contribution is simply given by multiplication between each coefficient and each feature value. In addition, the contributions need to be adjusted for the baseline values that \SHAP{} uses. We correct for these issues in our work for all considered explainers.
(2) No prior work has considered the case of feature interactions~\cite{miro2023novel,NEURIPS2020_ce758408,pmlr-v119-lakkaraju20a}, which are ubiquitous among black box models, especially neural networks. We consider models with a various number of feature interactions and order of feature interactions.
(3) Unlike prior work~\cite{miro2023novel,NEURIPS2020_ce758408,pmlr-v119-lakkaraju20a}, we consider in our experiments both synthetic and real-world data, both tabular and image data, as well as both non-learnable and learnable models, including convolutional neural networks.

Further comprehensive overviews of explanation evaluation methodologies and aspects
are detailed in~\cite{info14080469,anecdotalEvidenceXAI2023,zhouEvaluatingQualityMachine2021,Vermeire2022,pawelczyk2021carla,DBLP:journals/adac/RamonMPE20}.

\subsection*{Considered Local Post Hoc Explainers}
\begin{itemize}[noitemsep,nolistsep,leftmargin=*]
\item \textbf{\underline{P}artial \underline{D}ependence \underline{P}lots}~
\PDP{}s~\cite{Friedman2001} estimate the average marginal effect of a subset of features on the output of a model using the Monte Carlo method. When the subset comprises one or two features, the model output is plotted as a function of the feature values. \PDP{}{}s give a global understanding of a model, but can also yield a local explanation for the specific feature values of a sample.
\item \textbf{\underline{L}ocal \underline{I}nterpretable \underline{M}odel-agnostic \underline{E}xplanations}~
\LIME{}~\cite{ribeiroWhyShouldTrust2016} explains by learning a linear model from a randomly sampled neighborhood around z-score normalized instances. Feature selection is controlled by hyperparameters that limit the total number of features used in approximation, such as the top-$k$ largest-magnitude coefficients from a ridge regression model.
\item \textbf{\underline{M}odel \underline{A}gnostic Su\underline{p}ervised \underline{L}ocal \underline{E}xplanations}~
\MAPLE{}~\cite{plumbModelAgnosticSupervised2018} employs a tree ensemble, \eg{}, a random forest, to estimate the importance (the net impurity) of each feature. Feature selection is performed upfront on the background data by iteratively adding important features to a linear model until error is minimized on held out validation data. For local explanations, \MAPLE{} learns a ridge regression model on the background data distribution with samples weighed by the tree leafs relevant to the explained instance.
\item \textbf{\underline{Sh}apley \underline{A}dditive Ex\underline{p}lanations}~
\SHAP{}~\cite{lundbergUnifiedApproachInterpreting2017} takes a similar but distinct approach from \LIME{} by approximating the Shapley values of the conditional expectation of a model.
Feature selection is controlled using a regularization term.
Note that when we write \SHAP{}, we are specifically referring to Kernel \SHAP{}, which is distinguished from its other variants for trees,
structured data, etc.
An extension of \SHAP{} to handle dependent features has also been proposed~\cite{aasExplainingIndividualPredictions2020}, which we refer to as \SHAPR{} after the associated \texttt{R} package. In effort to improve the accuracy of \SHAP{} explanations, \SHAPR{} estimates the conditional distribution assuming features are statistically dependent.
\end{itemize}

\subsection*{\textsc{MatchEffects}}

\begin{figure*}
    \begin{center}
      \begin{subfigure}{0.95\linewidth}
        \includegraphics[width=\linewidth]{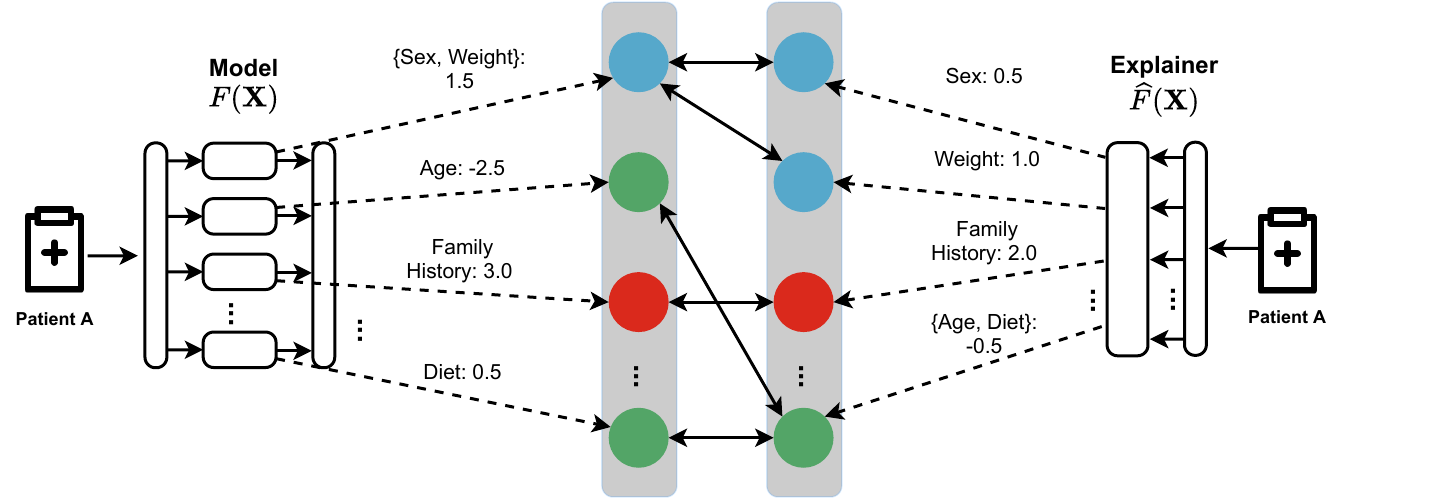}
        \caption{} \label{fig:matcheffectsex}
      \end{subfigure}\\%
      \begin{subfigure}{0.23\linewidth}
        \includegraphics[width=\linewidth]{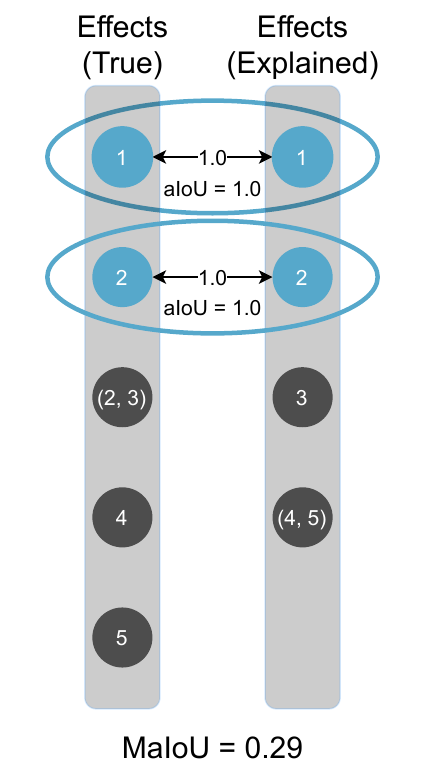}
        \caption{} \label{fig:matcheffectsa}
      \end{subfigure}%
      \begin{subfigure}{0.23\linewidth}
        \includegraphics[width=\linewidth]{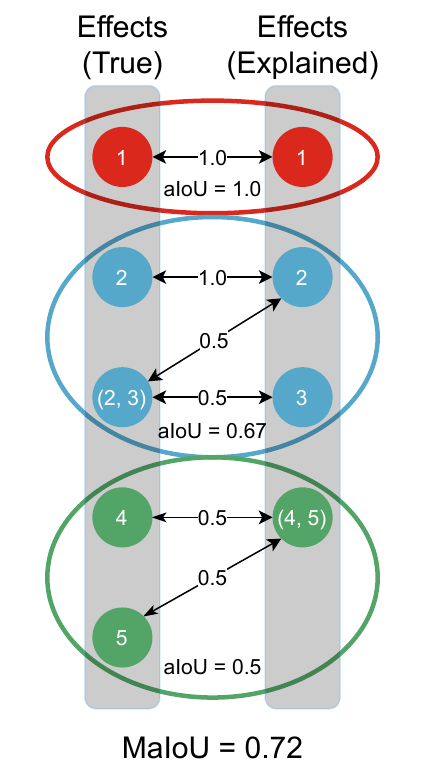}
        \caption{} \label{fig:matcheffectsb}
      \end{subfigure}%
      \begin{subfigure}{0.23125\linewidth}
        \includegraphics[width=\linewidth]{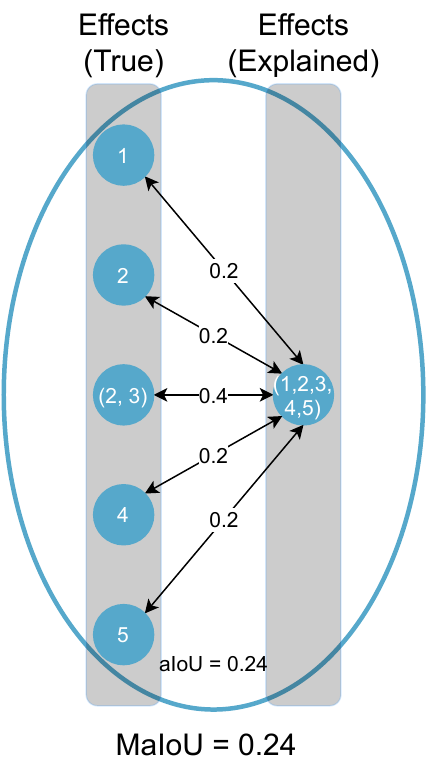}
        \caption{} \label{fig:matcheffectsc}
      \end{subfigure}
  \end{center}
    \caption{Examples of \textsc{MatchEffects} and MaIoU (Equation~\ref{eq:maiou}) in facilitating fair comparison between feature-additive explanations and \textit{ground truth}. (a) A simple example demonstrating \textsc{MatchEffects} for a hypothetical medical task. Like colors from each side of the graph can be directly compared. While no explainer evaluated in this work explicitly detects interactions ($|\hat{m}| \ge 2$), the visual demonstrates how such explainers are compatible with the framework.
    (b) A strict one-to-one matching between effects severs partially correct effects from comparison and yields an over-penalizing MaIoU. (c) With \textsc{MatchEffects}, effects are fairly grouped together for comparison and a more reasonable MaIoU is given -- ideally, the sum of the true contributions is equivalent to the sum of the explained contributions in each group (component). (d) Importantly, MaIoU defines the goodness of a match, in this case indicating that a superficially perfect explanation with \textsc{MatchEffects} (the sum of feature contributions is equivalent within in each group) is uninformative and incorrect.}
    \label{fig:matcheffects}
\end{figure*}

\textsc{MatchEffects} is formalized in Algorithm~\ref{alg:paireffects} and illustrated for a few examples in \figref{fig:matcheffects}.
This process guarantees
the most fair and direct comparison
of explanations, and does not rely on
gradients,
sensitivity,
or other proxy means~\cite{guidottiEvaluatingLocalExplanation2021,deyoungERASER2020,fengWhatCanAI2019}.
The worst-case time complexity of \textsc{MatchEffects} is
$\mathcal{O}\left(m \hat{m}d\right)$
and the
space complexity is
$\mathcal{O}\left(m \hat{m}\right)$
(see
Appendix C
for proofs).
It should be noted that in all practical use cases, the wall-time and memory bottlenecks of the framework arise from the explainers, especially those that scale combinatorially with $d$ or require a reference data set that scales with $n$.

\begin{algorithm}%
    \small
    \KwIn{$D = \{D_{j} \mid 1 \le j \le m_{F}\}$, the set of feature subsets operated on by
    model $F$}
    \KwIn{$\hat{D} = \{\hat{D}_{k} \mid 1 \le k \le m_{\hat{F}}\}$, the set of feature subsets operated on by
    explainer $\hat{F}$}
    \KwResult{Corresponding sets of effects that can be compared}
    \tcp{add edges between effects with mutual features}
    $E \gets \text{new array}$\;
    \For{$D_{j} \in D$}{
        \For{$\hat{D}_{k} \in \hat{D}$}{
            \If{$\lvert D_{j} \cap \hat{D}_{k} \rvert > 0$}{
                $E$.append($\{D_{j}, \hat{D}_{k}\}$)\;
            }
        }
    }
    $V \gets D \cup \hat{D}$\tcp*[r]{effects are vertices}
    $G \gets (V, E)$\;
    \tcp{find connected components ($CCs$) for the undirected graph $G$}
    $CCs \gets \textsc{ConnectedComponents}(G)$\;
    $\text{matches} \gets \text{new array}$\;
    \tcp{$V_c$ and $E_c$ comprise the vertices and edges of component $c$, respectively}
    \For(\tcp*[f]{unpack the components}){$\{V_c, E_c\} \in CCs$}{
        $\text{match}_F \gets \text{new array}$\;
        $\text{match}_{\hat{F}} \gets \text{new array}$\;
        \For{$D_c \in V_c$}{
            \eIf{$D_c \in D_{F}$}{
                $\text{match}_F$.append($D_c$)\;
            }{
                $\text{match}_{\hat{F}}$.append($D_c$)\;
            }
        }
        \If{$\textup{match}_F = \textup{match}_{\hat{F}}$}{
            \tcp{elements of identical sets are each a perfect match}
            \For{$D_c \in \textup{match}_F$}{
                $\text{matches}$.append($\{\{D_c\}, \{D_c\}\}$)\;
            }
        }{
            $\text{matches}$.append($\{\text{match}_F, \text{match}_{\hat{F}}\}$)\;
        }
        
    }
    \KwRet{$\textup{matches}$}
    \caption{\textsc{MatchEffects}}
    \label{alg:paireffects}
\end{algorithm}

One could exploit \textsc{MatchEffects} by producing explanations that attribute the entire output of the model to a single effect comprising all $d$ features; the comparison of contributions could trivially yield perfect but uninformative scores. Likewise, a model with such interaction effects, like most deep NNs, would render this evaluation inconsequential.
To mitigate this issue, we introduce a metric that evaluates the goodness of the matching.
Let $E_c$ be the set of edges of a single component found by \textsc{MatchEffects}. For an edge $\{D_{j}, \hat{D}_{k}\} \in E_c$, the intersection-over-union (IoU), also known as the Jaccard index, is calculated between $D_{j}$ and $\hat{D}_{k}$. The total goodness for a component is the average of the IoU scores of each edge in $E_c$, and the total goodness for a matching is the mean value of these averages: the mean-average-IoU (MaIoU). This metric is given by Equation~\eqref{eq:maiou}
\begin{align}\label{eq:maiou}
    \text{MaIoU}(CCs) &= \frac{1}{|CCs|} \sum_{\{V_c, E_c\} \in CCs} \text{aIoU}(E_c)
\intertext{%
and the average IoU (aIoU) is defined by Equation~\eqref{eq:aiou}%
}%
\label{eq:aiou}
    \text{aIoU}(E_c) &= \frac{1}{|E_c|} \sum_{\{D_{j}, \hat{D}_{k}\} \in E_c} \frac{|D_{j}\cap \hat{D}_{k}|}{|D_{j} \cup \hat{D}_{k}|}
    .
\end{align}
where $CCs$ is defined in Algorithm~\ref{alg:paireffects}.
MaIoU can be thought of the degree to which the effects uncovered by an explainer agree with the true effects of the model. \figref{fig:matcheffectsb} shows the effectiveness of MaIoU on an example with three components, and \figref{fig:matcheffectsc} shows how MaIoU can inform when an explanation is uninformative and mechanistically incorrect, mitigating the aforementioned consequences.

Table~\ref{tab:maiou} shows the MaIoU for each explainer on each dataset. Note that MaIoU is identical for both GAMs and NNs due to the experimental design: both are constrained to use the same (but still randomly selected) effects for each dataset. Of the explainers, \MAPLE{} has the worst (lowest) average MaIoU due to its feature selection process that picks relatively few features compared to the other explainers. \LIME{}, \SHAP{}, and \SHAPR{} achieve the same scores as they all provide feature-wise explanations and feature selection is not forced. This favors explanation completeness over human-comprehensibility, which is more favorable for testing fidelity. \PDP{} follows the same line of reasoning except for FICO; \PDP{} provides non-zero estimates of several more features than \LIME{} and \SHAP{}.
On the MNIST task, all explainers achieve the same MaIoU -- we explain this phenomenon in Appendix~D following the details of the feature-additive CNNs.

\begin{table}
    \centering
    \ra{1.1}
    \begin{tabular}{@{}lcccccc@{}}
        \toprule
        \multirow{2}{*}{Dataset}
        && \multicolumn{5}{c}{MaIoU} \\
        \cmidrule{3-7}
        && {\PDP{}} & {\LIME{}} & {\MAPLE{}} & {\SHAP{}} & {\SHAPR{}} \\
        \midrule
        \multirow{1}{*}{Boston}
            && 0.979&0.979&0.214&0.979&0.979 \\
        \multirow{1}{*}{COMPAS}
            && 0.971&0.971&0.286&0.971 & -- \\
        \multirow{1}{*}{FICO}
            && 0.750&0.659&0.648&0.659 & -- \\
        \multirow{1}{*}{MNIST}
            && 0.500&0.500&0.500&0.500&0.500 \\
        \bottomrule
    \end{tabular}
    \caption{The MaIoU for each explainer on several real-world datasets. Note that MaIoU is identical for both GAMs and NNs due to the experimental design: both are constrained to use the same (but still randomly selected) effects for each dataset. \SHAPR{} is not implemented for data with categorical variables in this work.}
    \label{tab:maiou}
\end{table}

\section{Reproducibility}
\label{sec:a}

\subsection*{Implementation and Setup}
We use
{\small\texttt{SymPy}}~\cite{sympy}
to generate synthetic models and represent expressions symbolically as expression trees. This allows us to automatically discover the additivity of arbitrary expressions.
See
Appendix C
for the unary and binary operators, parameters, and operation weights considered in random model generation.
All stochasticity is seeded for reproducibility, and all code is documented and open-sourced%
\footnote{The source code for this work is available at
\href{https://github.com/craymichael/PostHocExplainerEvaluation}{github.com/craymichael/PostHocExplainerEvaluation}%
}.
The framework is implemented in 
{\small\texttt{Python}}~\cite{van1995python}
with the help of {\small\texttt{SymPy}} and the additional libraries
{\small\texttt{NumPy}}~\cite{2020NumPy-Array},
{\small\texttt{SciPy}}~\cite{2020SciPy-NMeth},
{\small\texttt{pandas}}~\cite{pandas},
{\small\texttt{Scikit-learn}}~\cite{sklearn},
{\small\texttt{Joblib}}~\cite{joblib},
{\small\texttt{mpmath}}~\cite{mpmath},
{\small\texttt{pyGAM}}~\cite{pygam},
{\small\texttt{PDPbox}}~\cite{PDPbox},
{\small\texttt{alibi}}~\cite{alibi},
{\small\texttt{TensorFlow}}~\cite{tensorflow2015-whitepaper},
{\small\texttt{Matplotlib}}~\cite{matplotlib},
and
{\small\texttt{seaborn}}~\cite{seaborn}.
Furthermore, we build a {\small\texttt{Python}} interface to the
{\small\texttt{R}}~\cite{R}
package
{\small\texttt{shapr}}~\cite{shapr}
using
{\small\texttt{rpy2}}~\cite{rpy2}.
Appendix B
also details hyperparameters used to train the GAMs and feature-additive NNs, and the hardware used to run experiments.
Last, we consider the explanation of an effect to be $0$ if every estimated contribution is within a tolerance\footnote{See the documentation of \href{https://numpy.org/doc/1.19/reference/generated/numpy.allclose.html}{\texttt{numpy.allclose}} for details}. This is a fairer evaluation and tends to favor the explainers in experiments when dummy variables are present.

\subsection*{Model Generation Parameters}
See Algorithm~\ref{alg:synthetic} for definitions of parameters. The * in Table~\ref{tab:model_gen_params} means `1' is implied when $pct_{interact}$ is zero. 

\begin{table}[h]
    \centering
    \begin{tabular}{@{}cc@{}}
        \toprule
        Parameter & Values \\
        \midrule
        $d$ & \{2,    4,    7,   16,   32,   64,  127,  256,  512, 1024\} \\
        $n_{dummy}$ & \{0, 0.2375$d$, 0.475$d$, 0.7125$d$, 0.95$d$\} \\
        $pct_{nonlinear}$ & \{0, .375, .75, 1.125, 1.5\} \\
        $pct_{interact}$ & \{0, 0.167, 0.333, 0.5\} \\
        $order_{interact}$ & \{1*, 2, 3\} \\
        \bottomrule
    \end{tabular}
    \caption{Model generation parameters}
    \label{tab:model_gen_params}
\end{table}

Weights are the probability of being drawn as an operator (normalized against all considered operators in the considered classes). The add operation is only considered when the operator does not break up the interaction effect. The values of $n_{dummy}$ are a function of $d$.

\begin{table}[h]
    \centering
    \begin{tabular}{@{}rccc@{}}
        \toprule
        Name & Type & Nonlinear & Weight \\
        \midrule
        cosh$(\cdot)$ & unary & yes & 0.015 \\
        cosh$(\cdot)$ & unary & yes & 0.015 \\
        sin$(\cdot)$ & unary & yes & 0.015 \\
        sinh$(\cdot)$ & unary & yes & 0.015 \\
        asinh$(\cdot)$ & unary & yes & 0.015 \\
        tan$(\cdot)$ & unary & yes & 0.015 \\
        tanh$(\cdot)$ & unary & yes & 0.015 \\
        atan$(\cdot)$ & unary & yes & 0.015 \\
        cot$(\cdot)$ & unary & yes & 0.015 \\
        acot$(\cdot)$ & unary & yes & 0.015 \\
        csc$(\cdot)$ & unary & yes & 0.015 \\
        sech$(\cdot)$ & unary & yes & 0.015 \\
        sinc$(\cdot)$ & unary & yes & 0.015 \\
        $|\cdot|$ & unary & yes & 0.133 \\
        $\sqrt{(\cdot)}$ & unary & yes & 0.133 \\
        $(\cdot)^2$ & unary & yes & 0.133 \\
        $(\cdot)^3$ & unary & yes & 0.133 \\
        $\exp(\cdot)$ & unary & yes & 0.133 \\
        $\log(\cdot)$ & unary & yes & 0.133 \\
        \midrule
        $(\cdot) \times (\cdot)$ & binary & no & 0.8 \\
        $\sfrac{(\cdot)}{(\cdot)}$ & binary & no & 0.2 \\
        ${(\cdot)} + {(\cdot)}$ & binary & no & - \\
        \midrule
        $\min{(\cdot, \cdot)}$ & binary & yes & 0.5 \\
        $\max{(\cdot, \cdot)}$ & binary & yes & 0.5 \\
        \bottomrule
    \end{tabular}
    \caption{Operators considered}
    \label{tab:operators}
\end{table}

\subsection*{Explainer Hyperparameters}

In general, the defaults were used and no tuning was performed. We only allowed explainers to explain as many effects as possible as the goal wasn't to produce comprehensible explanations, but rather faithful ones as the only criteria. See the table below for specified parameters of interest. Note that we do not use L1 regularization with \SHAP{} as far too many features would be filtered and we do not tune explainer hyperparameters.

\begin{table}[h]
    \centering
    \begin{tabular}{@{}ccc@{}}
        \toprule
        Explainer &  \\
        \midrule
        \multirow{4}{*}{\LIME{}} & \texttt{num\_samples} & \texttt{5000} \\
         & \texttt{num\_features} & $d$ \\
         & \texttt{discretize\_continuous} & \texttt{False} \\
         & \texttt{feature\_selection} & \texttt{`auto'} \\
        \midrule
        \multirow{6}{*}{\MAPLE{}} & \texttt{train\_size} & \texttt{2/3} \\
         & \texttt{fe\_type} & \texttt{`rf'} \\
         & \texttt{n\_estimators} & \texttt{200} \\
         & \texttt{max\_features} & \texttt{0.5} \\
         & \texttt{min\_samples\_leaf} & \texttt{10} \\
         & \texttt{regularization} & \texttt{1e-3} \\
        \midrule
        \multirow{3}{*}{\SHAP{}} & \texttt{n\_background\_samples} & \texttt{100} \\
         & \texttt{summarization} & \texttt{kmeans} \\
         & \texttt{l1\_reg} & \texttt{False} \\
        \bottomrule
    \end{tabular}
    \caption{Explainer hyperparameters}
    \label{tab:explainer_hparams}
\end{table}

\subsection*{\PDP{} Local Explanations}
To generate local explanations using \PDP{}, we compute the \PDP{} for each feature individually. The \texttt{\PDP{}Box} library uses percentiles to sample the domain of each feature. We compute PD for 100 sample points for each feature. Thereafter, we use linear interpolation between each point, and extrapolation for values outside the range, to give a feature contribution for ``unseen'' values. The local explanation is thus the interpolated PD of each feature.

\subsection*{Dataset Descriptions}

We demonstrate our framework on 2,000 synthetic models that are generated with a varied number of effects, order of interaction, number of features, degree of nonlinearity, and number of unused (dummy) variables. For each, we discard models with invalid ranges and domains that do not intersect with the interval $[-1, 1]$. The data of each feature $\mathbf{x}_{*,i}$ is sampled independently from a uniform distribution $\mathcal{U}(-1, 1)$. We draw $n$ samples quadratically proportional to the number of features $d$ as $n = 500 \sqrt{d}$.
The explainers are evaluated on each model with access to the full dataset and black box access to the model.
Of the 2,000 models, 16 were discarded due to the input domain producing non-real numbers. Furthermore, some explainers were not able to explain every model due to invalid perturbations and resource exhaustion\footnote{See Appendix B for hardware and time budgets.}.
The former occurred with \PDP{}, \LIME{}, and \SHAP{}, typically due to models with
narrower feature domains, while the latter occurred with \MAPLE{} and \SHAPR{} due to
the inefficient use of background data and intrinsic computational complexity.
In total, 82\%, 39\%, 80\%, 91\%, and 40\% of models were successfully explained by \PDP{}, \LIME{}, \MAPLE{}, \SHAP{}, and \SHAPR{}, respectively.
We consider the failure to produce an explanation for valid input to be a limitation of an explainer or its implementation.

The Boston housing dataset~\cite{HARRISON197881} contains median home values in Boston, MA, that can be predicted by several covariates, including sensitive attributes, \eg{}, those related to race. Models that discriminate based off of such features necessitate that their operation be exposed by explanations.

We also evaluate explainers on the Correctional Offender Management Profiling for Alternative Sanctions~(COMPAS) recidivism risk dataset~\cite{angwin2016machine}.
The dataset was collected by ProPublica in 2016 and contains
covariates, such as criminal history and demographics, the proprietary COMPAS risk score, and recidivism data
for defendants from Broward County, Florida.

The FICO Home Equity Line of Credit~(HELOC) dataset~\cite{ficoheloc2018}, introduced in a 2018 XAI challenge, is also used in this work. It comprises anonymous HELOC applications made by consumers requesting a credit line in the range of \$5,000 and \$150,000. Given the credit history and characteristics of an applicant, the task is to predict whether they will be able to repay their HELOC account within two years.

Last, we evaluate on a down-sampled version of the MNIST dataset~\cite{MNIST}. With the aim of reducing explainer runtime and improving comprehensibility of effect-wise results, we crop and then resize each handwritten digit in the dataset
to $12\times 10$ and only include a subset of the 10 digits.
We evaluate explainers on a down-sampled version of the MNIST dataset as mentioned in the text. Again, all steps taken here are to reduce explainer run-times and improve the comprehensibility of the results. We select a subset of classes to reduce the amount of data and the number of classes to explain. Specifically, we include only the four digits 0, 1, 5, and 8 due to separability between the data of each class. The crop was selected by observing the percentage of non-zero pixels that would be removed for all crop values, i.e., of the top and bottom rows, and the left and right columns. We remove about 1\% of all non-zero pixels by using a global crop of 3 pixels from the top, 2 from the bottom, 5 from the left, and 3 from the right. This crop changes each image size from $28\times 28$ to $23\times 20$. We then resize each handwritten digit in the dataset to $12\times 10$ using the \texttt{scikit-image} function \texttt{resize} with anti-aliasing.

\subsection*{Experiment Reproducibility}
While all experiments use random seeds, some results may not be completely reproducible due to the behavior of \texttt{SymPy} (see the discussion in issue \#20522\footnote{\url{https://github.com/sympy/sympy/issues/20522}}).
For instance, the model generation uses the \texttt{sympy.calculus.util.continuous\_domain} function to determine if generated models have valid input domains. This function randomly iterates through assumptions, and due to bugs, may not converge to the same result. Thus, we provide every \texttt{SymPy} model in the \texttt{pickle} format, and the generated data, and true contributions, and the explainer contributions in a \texttt{NumPy} format. The latter files should not suffer from reproducibility issues, but are provided to guarantee reproducibility.
These files are located in a shared Google Drive folder:
\href{https://drive.google.com/drive/folders/1cBDwi4JIXmAihOv9yfjqrLNsohM-CX5W?usp=sharing}{https://drive.google.com/drive/folders/1cBDwi4JIXmAihOv9yfjqrLNsohM-CX5W?usp=sharing}.

The source code is also linked in the main paper with the same seeds used in our experiments as the default arguments.

For the neural network, we train each pathway (for each effect) with 3 fully-connected layers [64, 64, 32] with the first two using a ReLU activation and the latter with no activation (identity function). The Adam optimizer is used with a learning rate of 1e-3 and early stopping with a patience of 100 based on the training loss, restoring the best weights at the end of training. The maximum number of epochs is 1,000.
For the GAM, a spline term is added with 25 splines for each main effect, and a tensor term with 10 splines per marginal term is used for interaction effects. The link function is logistic for classification and identity for regression.

The convolutional neural network (CNN) uses the same optimization hyperparameters but a slightly different architecture. The first layer is 2D convolution with a kernel size and stride size of (2, 1), \texttt{SAME} padding (i.e., with a stride of one, the filtered output shape is the same as the input shape), and 4 filters. This implies sparsity within the model, thus additive structure. A dense layer then gives the output for each interaction effect from the output of the convolutional layer; due to the kernel and strides sizes, the filter outputs will all comprise the nonlinear ReLU function of 2 features.

For the real-world experiments, data is normalized (z-score normalization) before training. Training uses the full dataset as generalization is not of interest --- rather, we only care if explainers can faithfully explain the model's predictions. Feature contributions and data are inverse normalization in all figures for better readability in terms of the underlying features.

\subsection*{Hardware}
Experiments were run on a cluster running the Univa Grid Engine (UGE) software. Each job was allocated 16 cores of an Intel(R) Xeon(R) CPU E5-2680 v3 @ 2.50GHz and 10~GiB of RAM (soft maximum) per explanation of a model. Note that by pooling resources, a set of explanation jobs can contend for and pool up to 128~GiB of RAM. The operating system in use was Red Hat Enterprise Linux Server release 7.9 (Maipo). For total time running synthetic experiments, \LIME{} took $\sim$6 hours for all explanations, \SHAP{} took $\sim$18 hours for all explanations, and \MAPLE{} exceeded a 2 week budget (although, each run was faster than \SHAP{} up until $d > 64$). \SHAPR{} exceeded the memory limits for several processes, as well as the time budget of 2 weeks. \PDP{} finished all jobs in 6 days.

\subsection*{Licenses}
The FICO HELOC dataset license is available at \url{https://community.fico.com/s/explainable-machine-learning-challenge?tabset-3158a=a4c37}.
MNIST is under the Creative Commons Attribution-Share Alike 3.0 license. For software, see the corresponding licenses of the cited libraries in the text. Our software is under the MIT License.

\section{Proofs and Derivations}
\label{sec:b}

\subsection*{Proof of \textsc{MatchEffects} Complexities}
The worst-case time complexity of \textsc{MatchEffects} is
$\mathcal{O}\left(m \hat{m}d\right)$
and the
space complexity is
$\mathcal{O}\left(\max\left(d(m + \hat{m}), m \hat{m}\right)\right)$
(note that we write
$\mathcal{O}\left(m \hat{m}\right)$
in the text for simplicity as the number of effects is almost always $\gg d$ in practical usage of the algorithm).
Here we prove these claims, starting with the time complexity.

Lines 2-5 perform the following number of set intersections in the worst-case:
\begin{align*}
    &\phantom{=~}\mathcal{O}(|D| |\hat{D}|) \\
    &=\mathcal{O}(m \hat{m})
\end{align*}
\noindent
Similarly, this is the worst-case number of edges $|E| = m \hat{m}$, which occurs if the bipartite graph is fully-connected (all effects relate to all other effects).
Each set intersection (linear with hash sets in the implementation) has the following worst-case time complexity:
\begin{align*}
    &\phantom{=~}\mathcal{O}\left(
        \max{\left(\left\{
            |D_{j}| \mid
            D_{j} \in (D \cup \hat{D})
        \right\}\right)}
    \right) \\
    &=\mathcal{O}\left(
        |D_j^{\max}|
    \right)\\
    &=\mathcal{O}\left(
        d
    \right)
\end{align*}
\noindent
In the absolute worst-case every subset has $d$ features in the effect. Thus, the time complexity of these lines is $\mathcal{O}\left(m \hat{m}d\right)$.

In line 6, we compute the union of feature subsets (they become the graph vertices).
\begin{align*}
    &\phantom{=~}\mathcal{O}\left(
        |V|
    \right)\\
    &=\mathcal{O}\left(
        |D| + |\hat{D}|
    \right)\\
    &=\mathcal{O}\left(
        m + \hat{m}
    \right)
\end{align*}
\noindent
Set union takes linear time.

Line 8 performs the well-known connected components algorithm using graph traversal (BFS/DFS). Thus this takes
\begin{align*}
    &\phantom{=~}\mathcal{O}(|V|+|E|)\\
    &=\mathcal{O}\left(
        m + \hat{m} + m \hat{m}
    \right)\\
    &=\mathcal{O}\left(
        m \hat{m}
    \right)
\end{align*}
\noindent
time.

Lines 10-22 will traverse through each vertex exactly once (the vertices comprising each $V_c$ are guaranteed to be unique). For each vertex, checking membership in a set takes ($\mathcal{O}(1)$) time for each check. Thus, we have
$\mathcal{O}\left(
        |V|
    \right)
    =\mathcal{O}\left(
        m + \hat{m}
    \right)$
for these lines. The match equality comparisons over all iterations in the loop will also take the same time per the guarantee each vertex is visited once.
    
Thus, the worst-case time complexity is
$\mathcal{O}\left(m \hat{m}d\right)$\hfill$\square$

\vspace{2ex}

Here we consider the space complexity.
The size of the graph is simply the size of the vertices and edges
$\mathcal{O}\left(
        |V| + |E|
    \right)
    =\mathcal{O}\left(
        m \hat{m}
    \right)$.
Note that the graph is represented in a sparse format (nonzero values only), though this doesn't reduce space in the dense worst-case scenario.

The other space to consider is from $\text{matches}$. This contains sets, each with two sets of effects (ground truth and explained). For efficiency, each effect is represented as an index, reducing the space required from $\mathcal{O}(d)$ to $\mathcal{O}(1)$. Thus $\text{matches}$ ($|V|$ effects total, no matter how large a single match is) takes $\mathcal{O}(m + \hat{m})$ space.

Last, the input data is the same size as $\text{matches}$, except effects are not represented as indices. Therefore, the input data ($D$ and $\hat{D}$) takes $\mathcal{O}(d(m + \hat{m}))$ space.

The total space required is:

$\mathcal{O}\left(\max\left(d(m + \hat{m}), m \hat{m}\right)\right)$\hfill$\square$

\subsection*{Derivation of Equivalence Relations}

\subsubsection*{\LIME{}}

For \LIME{}, we derive the unnormalized coefficients for use in producing contributions. \LIME{} uses z-score normalization ($(x_i - \mu_i) / \sigma_i$) on the input data before learning local linear regression models.
\begin{align*}
    \hat{F}(\mathbf{x}) &= \theta_0 + \sum_i^d \frac{x_i - \mu_i}{\sigma_i} \theta_i \\
    \hat{F}(\mathbf{x}) &= \theta_0 + \sum_i^d \left(\frac{x_i}{\sigma_i} - \frac{\mu_i}{\theta_i}\right) \\
    \hat{F}(\mathbf{x}) &= \left(\theta_0 - \frac{\mu_i}{\theta_i}\right) + \sum_i^d \frac{x_i}{\sigma_i}
\intertext{%
Thus we simply just need to scale the coefficients as follows (same as the main text)
}
    \theta_0' &= \theta_0 - \sum_i \frac{\mu_i \theta_i}{\sigma_i} \\
    \theta_i' &= \frac{\theta_i}{\sigma_i}\\
\intertext{%
where $\theta_0'$ is the adjusted bias term and each $\theta_i'$ is an adjusted coefficient.
}
\end{align*}

\subsubsection*{\SHAP{}}
\SHAP{} estimates the contributions relative to the mean-centered model response. In other words:
\begin{align*}
    \hat{F}(\mathbf{x}) &\approx F(\mathbf{x}) - \mathbb{E}\left[F(\mathbf{x})\right] \\
\intertext{%
Thus, the \SHAP{} estimation can be written as follows
}
    \hat{F}(\mathbf{x}) &= \sum^d_i \hat{f}_i\left(\mathbf{x}_{i}\right) - \mathbb{E}[f_i(\mathbf{x}_{i})]\\
\intertext{%
due to the fact that
}
    \mathbb{E}\left[F(\mathbf{x})\right] &= \mathbb{E}\left[\sum_j^m f_j\left(\mathbf{x}_{D_{j}}\right)\right]\\
    &= \sum_j^m \mathbb{E}\left[f_j\left(\mathbf{x}_{D_{j}}\right)\right].\\
\intertext{%
So for each contribution, we can write that of the explainer as
}
    \hat{C}_{i} &= \hat{f}_i(\mathbf{x}_{i}) + \mathbb{E}[C_{i}]\\
\intertext{%
in order to correct for the removed expected value.
However, this assumes that there is some $i = j = k$ for every $f_j(\cdot)$ and $\hat{f}_k(\cdot)$ of the white box and explainer. As this is not always the case, we have to consider the effects of a match holistically. This then gives us the final relation for some match:
}
    C_{\text{match}_{\hat{F}}} &= \sum_{k \in \text{match}_{\hat{F}}} \hat{f}_k(\mathbf{x}_{k}) + \sum_{j \in \text{match}_{F}} \mathbb{E}[C_{j}]
\end{align*}

\noindent
This same process applies to \SHAPR{}.

\section{Additional Results and Figures}
\label{sec:c}

\begin{figure*}%
    \centering
    \begin{subfigure}{.63\linewidth}
        \includegraphics[width=\linewidth]{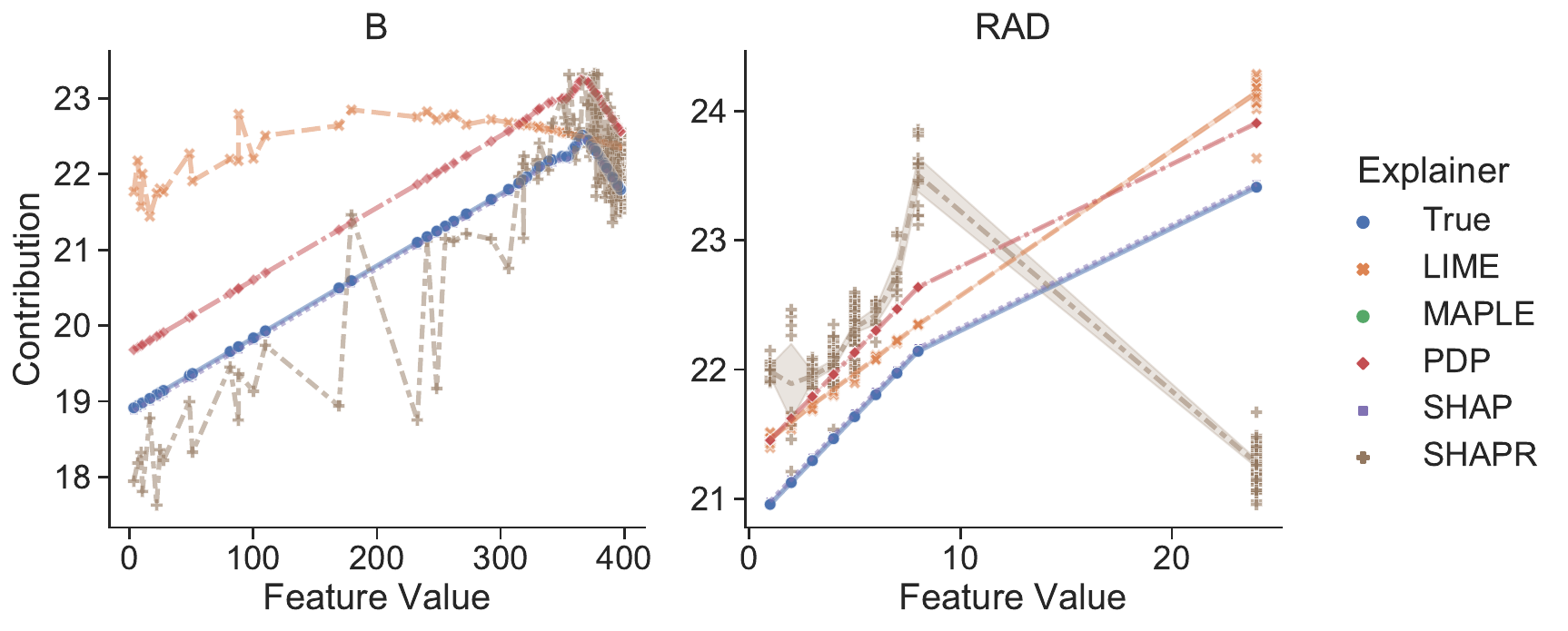}
        \caption{}
      \end{subfigure}%
      \begin{subfigure}{.355\linewidth}
        \includegraphics[width=\linewidth]{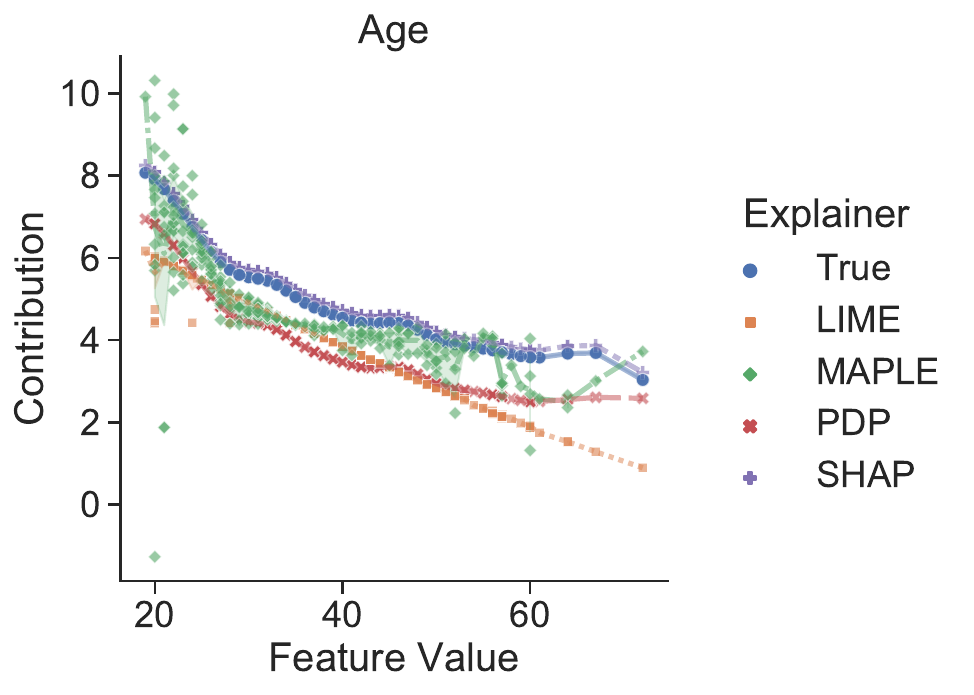}
        \caption{}
      \end{subfigure}%
    \caption{The true and explained feature shapes of (a) RAD (index of accessibility to radial highways) and B (proportion of African American population by town\protect\footnotemark) from a NN trained on the Boston housing dataset. \SHAPR{}, \MAPLE{}, and \LIME{} fail to satisfactorily unearth how the proportion of African Americans living in an area (feature B), drive the housing price. Fortunately, in this instance, \SHAP{} reveals this relationship within reasonable error. (b) The true and explained feature shapes of Age from a GAM trained on the COMPAS dataset. Several explainers produce misleading and noisy explanations, and some explained feature shapes correlate with the ground truth but are offset across all feature values. See Appendix~C for feature shapes of all remaining main effects on each task.}
    \label{fig:boston_subset}
\end{figure*}

We also visualize a subset of results in \figref{fig:boston_subset} as feature shapes.
We consider normalized (interquartile) root-mean-square error (\textsc{nrmse}) for comparing individual effects, as defined by Eq.~\eqref{eq:nrmse}
\begin{equation}\label{eq:nrmse}
    \text{\textsc{nrmse}}(\mathbf{a}, \mathbf{b}) = \frac{1}{Q_3^\mathbf{a} - Q_1^\mathbf{a}} \sqrt{\frac{\sum_i^n (a_i - b_i) ^ 2}{n}}
\end{equation}
\noindent
where $Q_3^\mathbf{a}$ and $Q_1^\mathbf{a}$ are the third and first quartiles of $\mathbf{a}$, respectively.
This measure of error is the quantified \textit{infidelity} of explanations.
This shows more clearly that several explainers do not faithfully explain some of the feature contributions.
For example, \SHAPR{}, \MAPLE{}, and \LIME{} fail to satisfactorily unearth how the proportion of African Americans living in an area (feature B), according to the NN, drive the housing price;
\SHAPR{} produces a high-variance estimate ($\text{\textsc{nrmse}} = 1.68$),
\MAPLE{} fails to even detect the effect ($\text{\textsc{nrmse}} = 3.59$),
and
\LIME{} only is able to approximate the mean contribution value ($\text{\textsc{nrmse}} = 3.04$).
This type of failure is incredibly misleading to any user and potentially damaging if the model is deployed.
Fortunately, in this instance, \SHAP{} reveals this relationship within reasonable error ($\text{\textsc{nrmse}} = 0.129$).
The COMPAS visualization shows another example of explanations of the ``Age'' feature of a GAM. Again, several explainers produce misleading and noisy explanations.
Notably, some explained feature shapes correlate with the ground truth (\eg{}, ``RAD'') but are offset (the expected value of the feature contributions deviate). In turn, this becomes a problem when the ranks of feature contributions are considered, which is how many interpretability tools present explanations~\cite{alibi,lundbergUnifiedApproachInterpreting2017,ribeiroWhyShouldTrust2016}.

\addtocounter{footnote}{1}

\footnotetext{While the Boston housing dataset is widely studied as a baseline regression problem, the data column (``B'') is notably controversial; the original paper~\cite{HARRISON197881} includes and preprocesses the data as $B = 1000(B' - 0.63)^2$ where $B'$ is the proportion of African Americans by town.}

\begin{figure}
    \centering
    \includegraphics[width=.6\linewidth]{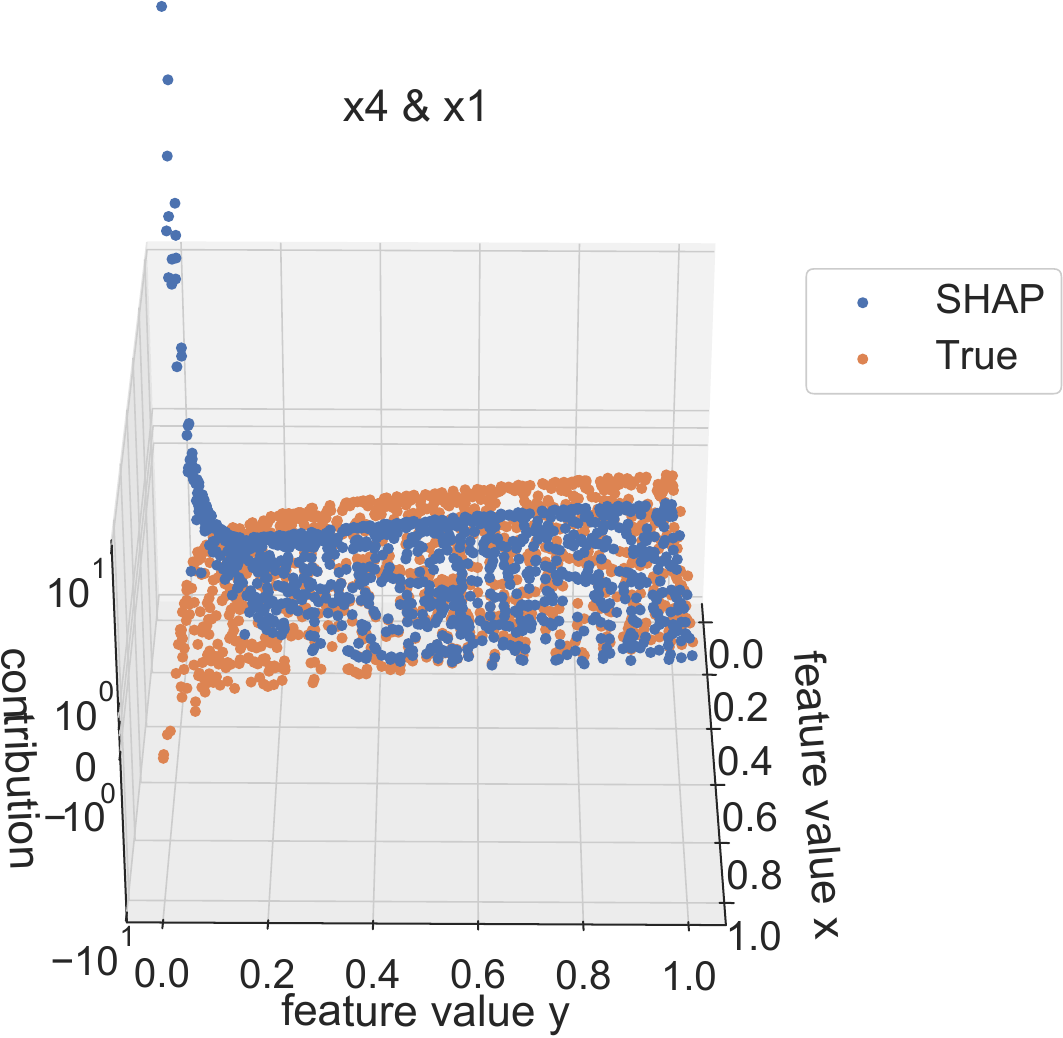}
    \caption{\SHAP{} explanations for the generated synthetic expression: $x_{1} + e^{x_{4}} + \log{\left(x_{1} x_{4} \right)} + \frac{x_{4}}{x_{1}}$. The expression is an interaction effect of the two variables $x_{1}$ and $x_{4}$, so the application of \textsc{MatchEffects} results in the comparison of the sum of the explained contributions of each variable. As feature value y approaches 0, the \SHAP{}-estimated contributions deviate exponentially from the ground truth. See Appendix C for additional angles and examples of explanations.}
    \label{fig:shap_bad_3d}
\end{figure}

Surprisingly, \SHAPR{} struggles to handle interaction effects effectively -- the baseline \SHAP{} outperforms it substantially.
While the quantitative comparison is clear, it is difficult to intuit poor explanations. In turn, we visualize an instance of an explained interaction effect by the best-performing explainer, \SHAP{}, in \figref{fig:shap_bad_3d}. As the feature value on the $y$-axis decreases, \SHAP{} and the ground truth contributions deviate exponentially (average cosine distance of 0.492 and Euclidean distance of 2.54).

\begin{figure*}[t]
    \centering
    \begin{subfigure}{0.45\linewidth}
        \includegraphics[width=\linewidth]{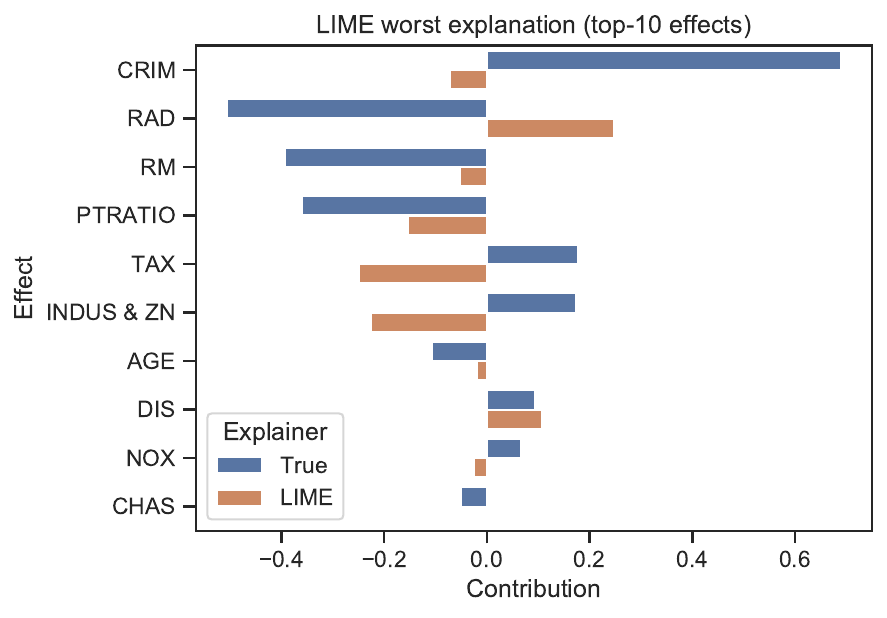}
    \end{subfigure}%
    \begin{subfigure}{0.45\linewidth}
        \includegraphics[width=\linewidth]{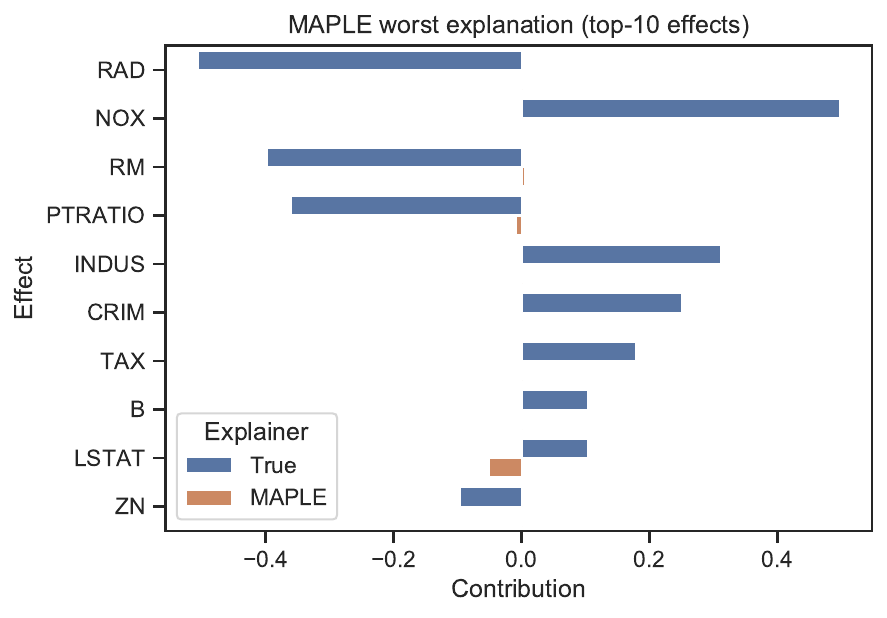}
    \end{subfigure}\\%
    \begin{subfigure}{0.45\linewidth}
        \includegraphics[width=\linewidth]{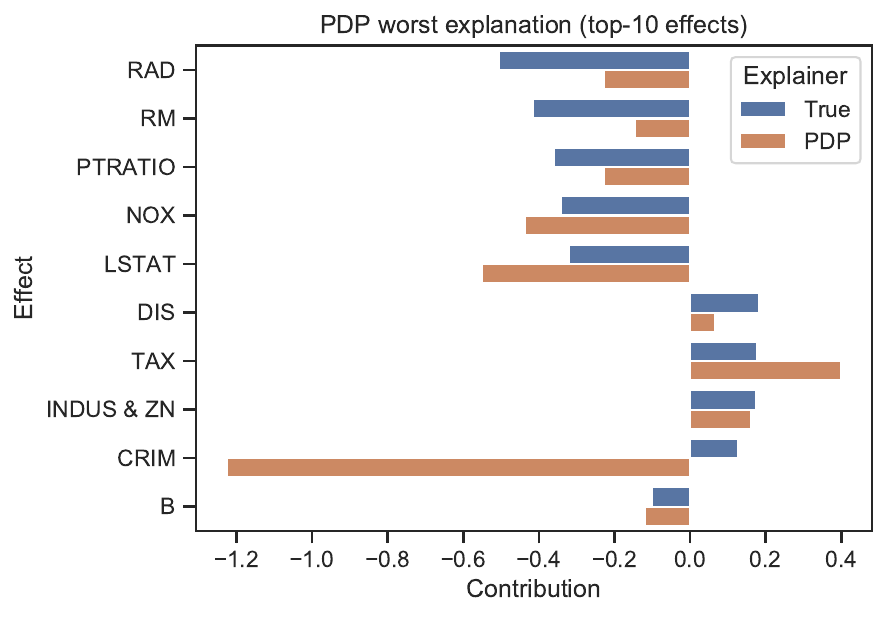}
    \end{subfigure}%
    \begin{subfigure}{0.45\linewidth}
        \includegraphics[width=\linewidth]{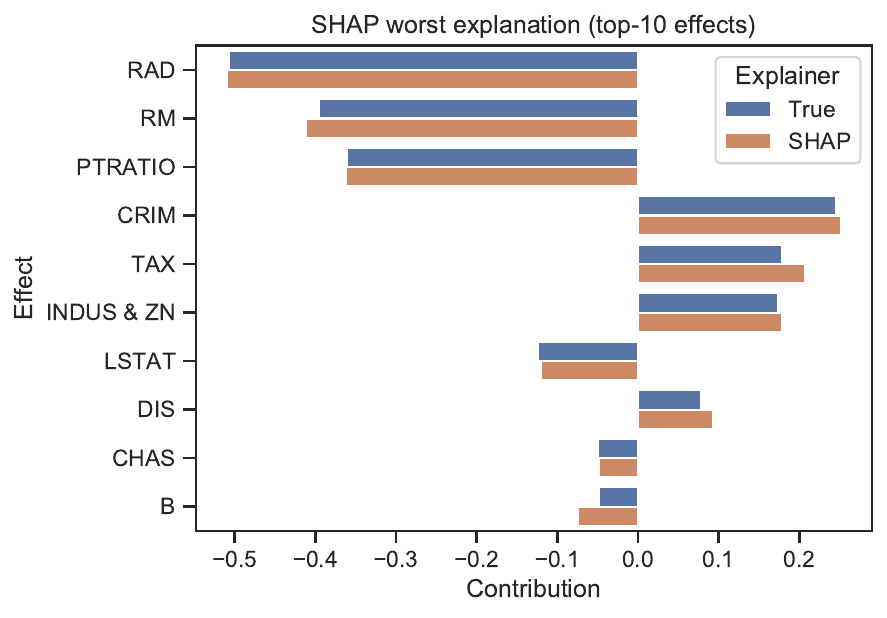}
    \end{subfigure}\\%
    \begin{subfigure}{0.45\linewidth}
        \includegraphics[width=\linewidth]{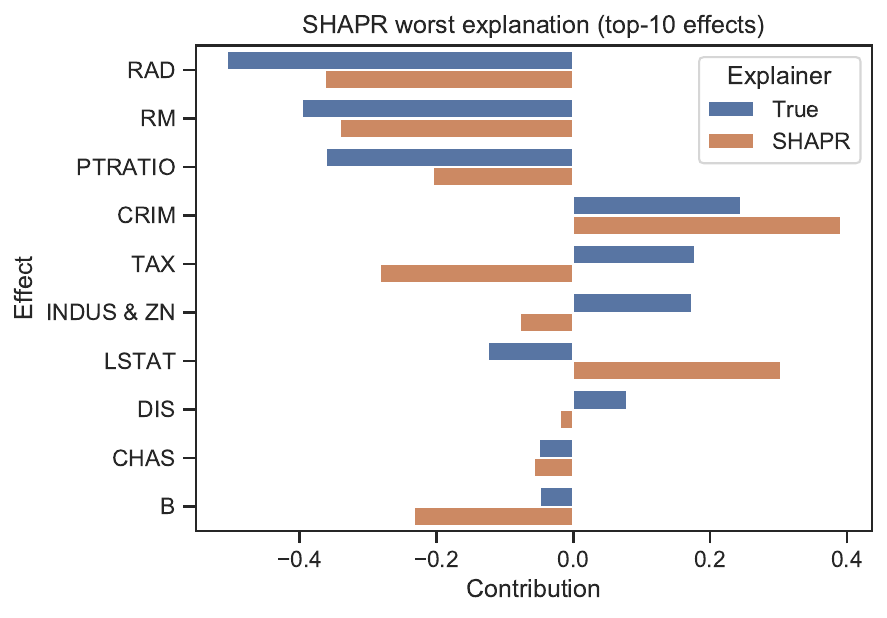}
    \end{subfigure}%
    \caption{The top-10 explained effects of the worst explanation of a GAM trained on the Boston dataset from each explainer. Top effects are ranked by magnitude and the quality of explanation is ranked by the mean cosine distance among all explained samples.}
    \label{fig:topk_worst_explanations}
\end{figure*}

\begin{figure*}[t]
    \centering
    \begin{subfigure}{0.4\linewidth}%
        \includegraphics[width=\linewidth]{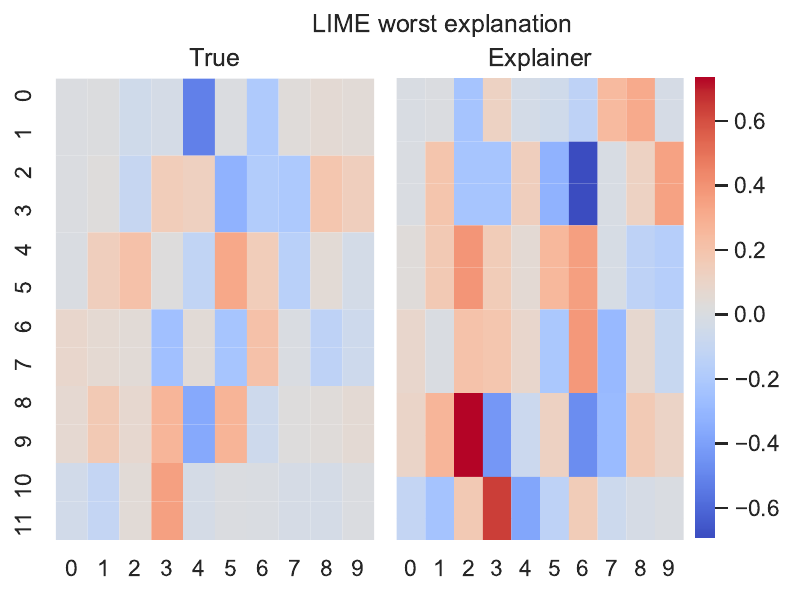}
    \end{subfigure}%
    \begin{subfigure}{0.4\linewidth}%
    \includegraphics[width=\linewidth]{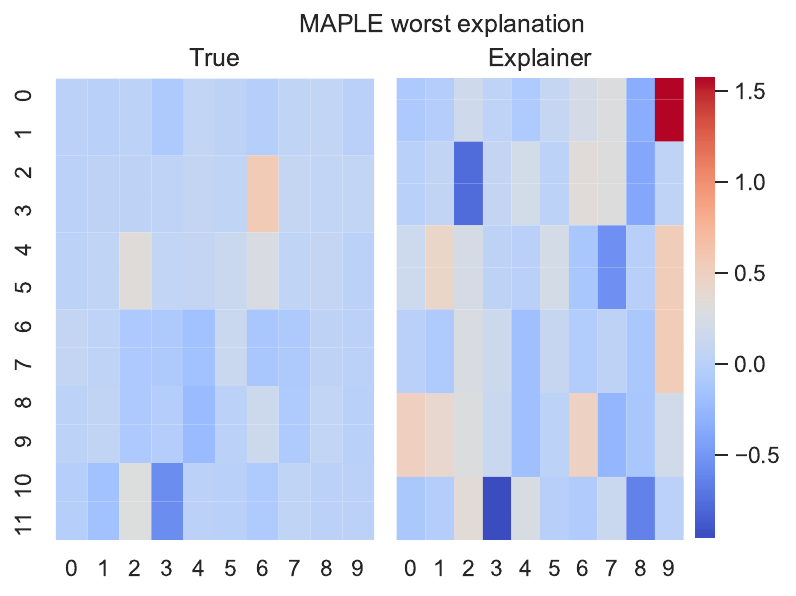}
    \end{subfigure}\\%
    \begin{subfigure}{0.4\linewidth}%
    \includegraphics[width=\linewidth]{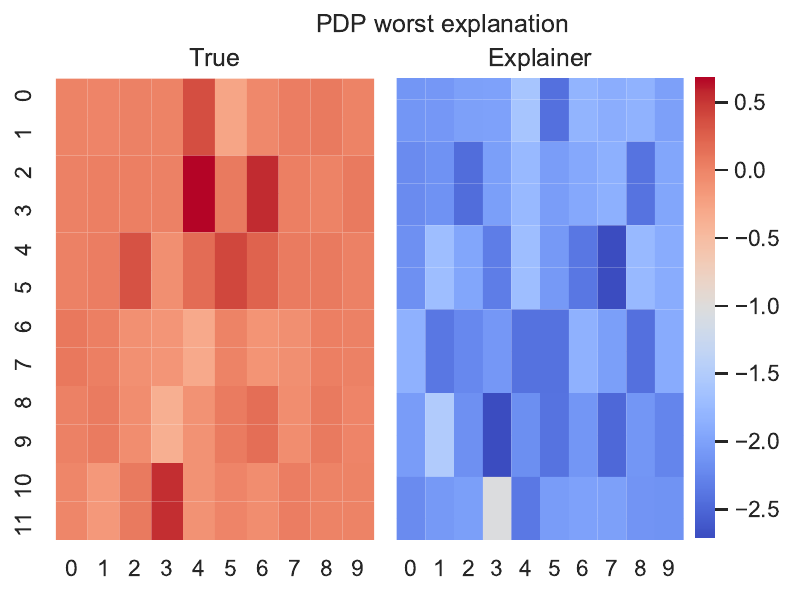}
    \end{subfigure}%
    \begin{subfigure}{0.4\linewidth}%
    \includegraphics[width=\linewidth]{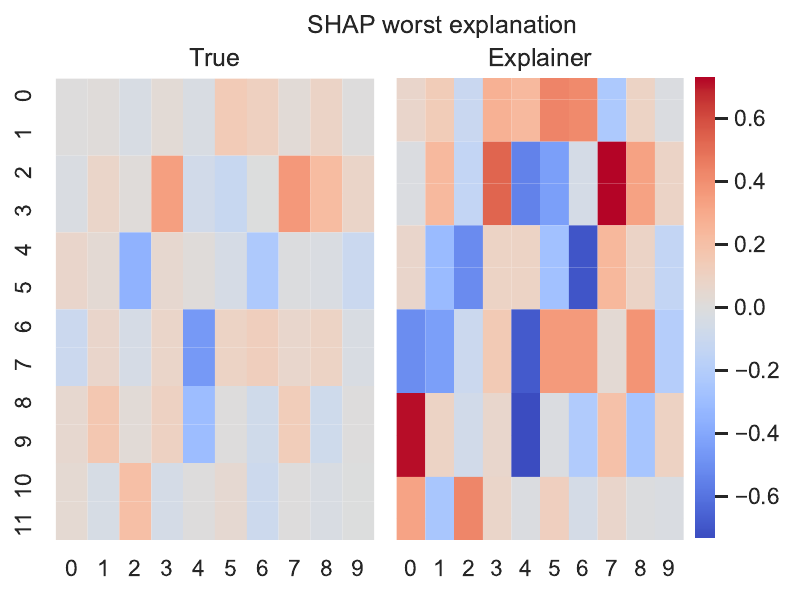}
    \end{subfigure}\\%
    \begin{subfigure}{0.4\linewidth}%
    \includegraphics[width=\linewidth]{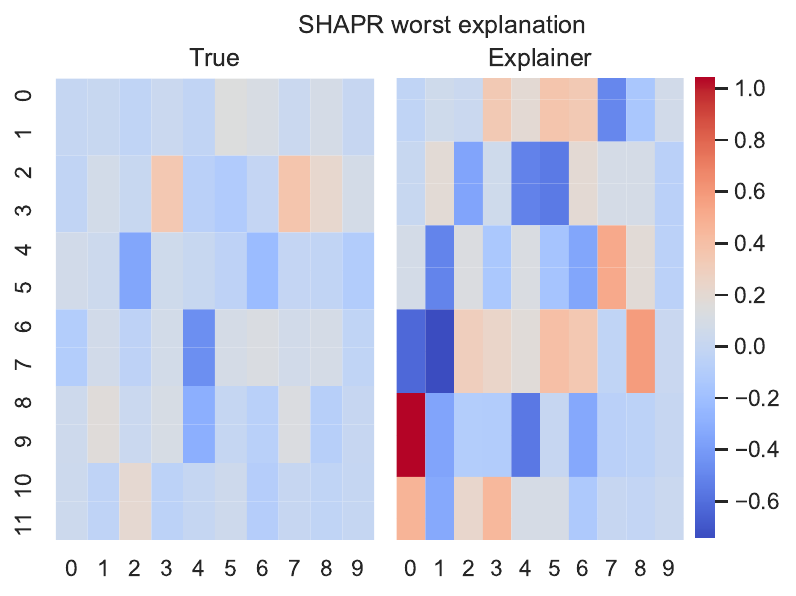}
    \end{subfigure}%
    \caption{The heatmap of explained interaction effects of the worst explanations by \LIME{}, \MAPLE{}, \PDP{}, \SHAP{}, and \SHAPR{} (left-to-right, top-to-bottom, respectively) for a CNN trained on the MNIST dataset as described in the text. See Appendix~C for details on the CNNs in experiments and the derived interaction effects. The worst explanation is determined by the cosine distance from the ground truth explanations.}
    \label{fig:worst_expls_mnist}
\end{figure*}

Looking at aggregate metrics is not sufficient to understand how poor an explanation may be. We consider the utility of an explainer in high-stakes applications to be limited by its worst explanation. Consequently, we visualize the worst explanations from each explainer and show those on the Boston dataset in \figref{fig:topk_worst_explanations}. The 10 most relevant effects to each decision are shown and the quality of the explanation is assessed using cosine distance. Again, the low point of \SHAP{} is still of relatively high quality, and the other explainers reveal incorrect attributions of effects. \MAPLE{} selects few important features correctly and does so conservatively. \PDP{}, \LIME{}, and \SHAPR{} all problematically assign opposite-signed contributions for several effects.
Similarly, we visualize the worst explanations for the MNIST task in \figref{fig:worst_expls_mnist}. Every explainer manages to flip the sign of at least a few contributions with the extreme case being \PDP{}. While \SHAP{} performs notably better than the other explainers, as shown in Table~\ref{tab:real_world_results}, its worst explanation is qualitatively misleading with some exaggerated contributions and some contributions with the opposite sign. The MNIST task of explaining the CNN is more difficult due to the higher dimensionality of the data and the number of interaction effects in the CNN. These interaction effects are of course due to the convolutional layers that operate over local neighborhoods of pixels. Appendix~A goes into greater detail on this point.

\section{Synthetic Model Generation}
\label{sec:d}

Synthetic models are generated as described by Algorithm~\ref{alg:synthetic}, \textsc{GenerateModel}.
This algorithm takes in three absolute parameters: the number of features, the number of dummy (unused) features, and the order of interactions. It also takes in two relative parameters: the percentage of nonlinear operators and the percentage of interaction terms.
Generation is split into four phases: nonlinear main effects, linear main effects, nonlinear interaction effects, and linear interaction effects. These phases are marked by corresponding comments in the algorithm.

Before any phase, we select the unique features to use in the model, which is simply the $d$ features with the dummy features removed from consideration. After model generation, data is still drawn for these unused variables, but the model ignores it.
For nonlinear main effects, we use at most the percentage of nonlinear operators times the number of features as the number of effects to generate. If this product is larger than the number of features, then we determine that the residual nonlinear operators will be applied to effects multiple times. For example, for $d=2$ and a nonlinear percentage of 2 (200\%), we may end up with something like $\cos(|x_1|) + \exp(\sqrt{x_2})$.
This describes the steps where we place operators into some amount of bins (which is done as uniformly as possible with the values of each bin being an integer).
Following this selection, we simply iterate over the unique features, applying the unary nonlinear operators to each, and add the result (still in symbolic form) to the expression.
For linear main effects, we simply add the number of remaining features that have not had nonlinearities applied, if any, to the expression.

We have two parameters to consider for interaction effects: the interaction order (the number of features involved in each interaction) and the percentage of interaction terms (treated in the same manner as the percentage of nonlinear operators).
We first select the unique interactions based on the number of interactions specified and the number of main effects. This is a simple way of constraining the sparsity of generated models --- with too many interaction terms, separation of effect contributions may not be possible by \textsc{MatchEffects}. The unique interactions selected are also naturally limited by the number of possible unique combinations given the number of features and the order of interactions. From these interactions, we select the number to be nonlinear in the exact same way as the main effects. However, we make choices from both unary and binary operators --- binary operators are used to bridge together terms to form a whole effect and can include linear binary operators if the number of nonlinear operators is not sufficient to do so (\textit{i.e.}, less than the number of features in an interaction minus one). Finally, we select the remaining linear interactions, choose linear interaction operators, and additionally add these to the expression.

See the previous supplemental content listing the unary and binary operators. The implementation of this algorithm has all randomness, \textit{e.g.}, choices, seeded. For simplicity, the data structures (binary expression trees), random choices with operator weights, and valid domain checking are omitted from this algorithm.

\begin{algorithm}[t]
    \small
    \KwIn{$d$: the number of features}
    \KwIn{$n_{dummy}$: the number of unused features}
    \KwIn{$pct_{nonlinear}$: the percentage of nonlinearities used (relative to $d$)}
    \KwIn{$pct_{interact}$: the percentage of interaction terms (relative to $d$)}
    \KwIn{$order_{interact}$: the order of interaction terms ($\ge 2$)}
    \KwResult{A randomly generated expression (model)}
    $features \gets \text{ choose } (d - n_{dummy}) \text{ unique features}$\;
    \tcp{Initialize Expression}
    $expr \gets 0$\;
    \tcp{Nonlinear Main Effects}
    $n'_{main\_nonlinear} \gets pct_{nonlinear} \times |features|$\;
    \tcp{Number of terms}
    $n_{main\_nonlinear} \gets \min(n'_{main\_nonlinear}, |features|)$\;
    \tcp{Each bin will on average contain $n'_{main\_nonlinear} / n_{main\_nonlinear}$ operators}
    $ops_{main\_nonlinear} \gets $ \text{ place } $n'_{main\_nonlinear}$ unary nonlinear operators into $n_{main\_nonlinear}$ \text{ bins}\;
    \tcp{Cycle keeps track of the current element in a sequence, starting at the beginning if the previous element was at the end}
    $main_{features} \gets \text{cycle}(features)$\;
    \For{$i \in \{i \mid 1 \le i \le n_{main\_nonlinear}$}{
        \tcp{Get the next feature in the cycle}
        $term \gets \text{ next } main_{features}$\;
        \tcp{Apply nonlinearities}
        \For{$op \in ops_{main\_nonlinear}[i]$}{
            $term \gets op(term)$\;
        }
        $expr \gets expr + term$\;
    }
    \tcp{Linear Main Effects}
    $n_{main\_linear} \gets |features| - n_{main\_nonlinear}$\;
    \For{$i \in \{i \mid 1 \le i \le n_{main\_linear}\}$}{
        $feature \gets \text{ next } main_{features}$\;
        $expr \gets expr + feature$\;
    }
    \tcp{Nonlinear Interaction Effects}
    $n_{interact} \gets \min{\{pct_{interact} \times |features|, |features|\}}$\;
    $n'_{interact\_nonlinear} \gets pct_{nonlinear} \times n_{interact}$\;
    $interactions \gets $ choose $n_{interact}$ unique feature pairs of size $order_{interact}$\;
    $n_{interact\_nonlinear} \gets \min(n'_{interact\_nonlinear}, n_{interact})$\;
    \tcp{Each is a unary/binary nonlinear operator or binary linear operator. \# of binary operators per effect $= order_{interact} - 1$}
    $ops_{interact\_nonlinear} \gets $ \text{ place } $n'_{interact\_nonlinear}$ (non)linear operators into $n_{interact\_nonlinear}$ \text{ bins}\;
    $interact_{features} \gets \text{cycle}(interactions)$\;
    \For{$i \in \{i \mid 1 \le i \le n_{interact\_nonlinear}$}{
        $interaction \gets \text{ cycle$($next } interact_{features})$\;
        $term \gets \text{ next } interaction$\;
        \For{$op \in ops_{interact\_nonlinear}[i]$}{
            \eIf{$op$ \textup{is unary}}{
                $term \gets op(term)$\;
            }{
                $feature \gets \text{ next } interaction$\;
                $term \gets op(term, feature)$\;
            }
        }
        $expr \gets expr + term$\;
    }
    \tcp{Continues on the following page...}
    \caption{\textsc{GenerateModel}: Generates a synthetic model satisfying various arguments}
    \label{alg:synthetic}
\end{algorithm}

\begin{algorithm}[t]
    \setcounter{AlgoLine}{31}
    \small
    \tcp{Linear Interaction Effects}
    $n_{interact\_linear} \gets n_{interact} - n_{interact\_nonlinear}$\;
    \For{$i \in \{i \mid 1 \le i \le n_{interact\_linear}\}$}{
        $interaction \gets \text{ cycle$($next } interact_{features})$\;
        $ops_{interact\_linear} \gets $ choose $|interaction| - 1$ linear non-additive binary operations\;
        $term \gets \text{ next } interaction$\;
        \For{$op \in ops_{interact\_linear}$}{
            $feature \gets \text{ next } interaction$\;
            $term \gets op(term, feature)$\;
        }
        $expr \gets expr + feature$\;
    }
    \KwRet{$expr$}
\end{algorithm}

\end{document}